\documentclass[11pt]{article}
\usepackage{fullpage}
\usepackage[final]{hyperref}
\usepackage{nameref}
\usepackage[round]{natbib}
\usepackage{tabularx}
\usepackage{booktabs}
\usepackage{setspace}
\usepackage{makecell}
\usepackage{graphicx}
\usepackage{wrapfig}
\usepackage{subcaption}
\usepackage{multirow}
\usepackage{delarray}
\usepackage{xcolor}
\usepackage{colortbl}
\usepackage{amssymb}
\usepackage{amsmath}
\usepackage{mathrsfs}
\usepackage{mathtools}
\usepackage{nicefrac}       
\usepackage{tikz}
\usepackage{xspace}
\xspaceaddexceptions{[]\{\}}
\usepackage{algorithm}
\usepackage{algorithmic}
\usepackage{mdframed}
\usepackage{bm}
\usepackage{tablefootnote}
\usepackage{threeparttable}
\allowdisplaybreaks

\usepackage{tabularray}
\usepackage{placeins}	

\usepackage[capitalize]{cleveref} 

\newtheorem{assumption}{Assumption}
\newtheorem{theorem}{Theorem}
\newtheorem{lemma}{Lemma}

{\hspace*{\fill}$\Box$\par}
{\hspace*{\fill}$\Box$\par\vspace{4mm}}
{\hspace*{\fill}$\Box$\par}

\hypersetup{
	colorlinks=true,       
	linkcolor=blue,        
	citecolor=blue,        
	filecolor=magenta,     
	urlcolor=blue
}

\newcommand*{\qedb}{\hfill\ensuremath{\square}}   

\newcommand{\topic}[1]{\vspace{2mm}\noindent{{\bf #1.}}}

\newcommand{\E}{{\mathbb{E}}}
\newcommand{\R}{{\mathbb{R}}}

\newcommand{\inner}[2]{\langle #1,#2 \rangle}

\newcommand{\ns}[1]{\| #1 \|^2}

\newcommand{\n}[1]{\| #1 \|}

\newcommand{\newll}{\notag \\ &\qquad \qquad }

\newcommand{\dataset}[1]{\texttt{#1}\xspace}

\newcommand{\XVFL}{{\sf \small X-VFL}\xspace}
\newcommand{\XCom}{{\sf \small XCom}\xspace}
\newcommand{\DSAlign}{{\sf \small DS-Align}\xspace}

\newcommand{\eat}[1]{}

\begin{document}

\title{\Large \bf \textsf{X-VFL}: A New Vertical Federated Learning Framework with Cross Completion and Decision Subspace Alignment}
\author{Qinghua Yao\thanks{Equal contribution. The work of Qinghua Yao and Xiangrui Xu was conducted during their time as Visiting Research Students in Zhize Li's group at SMU.} \\
	Singapore Management University\\ 
	\& University of Pennsylvania \\
	\and
	Xiangrui Xu\footnotemark[1]  \\
	Singapore Management University \\ 
	\& Beijing Jiaotong University\\
	\and
	Zhize Li\thanks{Corresponding author (\texttt{zhizeli@smu.edu.sg}).}\\
	Singapore Management University \\
	}

\date{August 8, 2025}
\maketitle

\begin{abstract}
	Vertical Federated Learning (VFL) enables collaborative learning by integrating disjoint feature subsets from multiple clients/parties. However, VFL typically faces two key challenges: i) the requirement for perfectly aligned data samples across all clients (missing features are not allowed); ii) the requirement for joint collaborative inference/prediction involving all clients (it does not support locally independent inference on a single client). To address these challenges, we propose \XVFL, a new VFL framework designed to deal with the non-aligned data samples with (partially) missing features and to support locally independent inference of new data samples for each client. In particular, we design two novel modules in \XVFL: \emph{Cross Completion} (\XCom) and \emph{Decision Subspace Alignment} (\DSAlign). \XCom can complete/reconstruct missing features for non-aligned data samples by leveraging information from other clients. \DSAlign aligns local features with completed and global features across all clients within the decision subspace, thus enabling locally independent inference at each client. Moreover, we provide convergence theorems for different algorithms used in training \XVFL, showing an $O(1/\sqrt{T})$ convergence rate for SGD-type algorithms and an $O(1/T)$ rate for PAGE-type algorithms, where $T$ denotes the number of training update steps. Extensive experiments on real-world datasets demonstrate that \XVFL significantly outperforms existing methods, e.g., achieving a 15\% improvement in accuracy on the image \dataset{CIFAR-10} dataset  and a 43\% improvement on the medical \dataset{MIMIC-III} dataset. These results validate the practical effectiveness and superiority of \XVFL, particularly in scenarios involving partially missing features and locally independent inference.
\end{abstract}

\section{Introduction} \label{sec:intro}

Federated Learning (FL) is a collaboratively learning framework where multiple clients/parties jointly train a machine learning model under the coordination of a central server without sharing their raw data~\citep{first_paper, HFL_survey1}.
Based on the partitioning of sample and feature spaces, FL can be categorized into Horizontal FL (HFL) and Vertical FL (VFL)~\citep{VFL3, HFL_survey1, VFL_survey}. 
In HFL, different clients typically hold different local datasets that share the same feature space, whereas in VFL, all clients use the same set of data samples but each holds a different subset of features.

\begin{wrapfigure}{r}{0.5\textwidth}
	\centering
	\includegraphics[width=\linewidth]{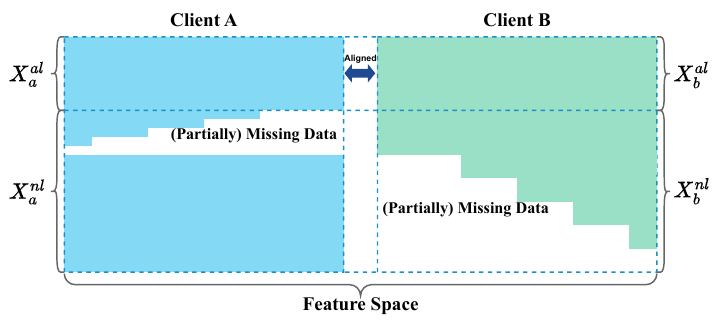}
	\vspace{-5mm}
	\caption{Dataset partition in VFL with $k=2$.}
	\label{fig:datasetting}
\end{wrapfigure}
In this paper, we focus on VFL, which is particularly useful in domains such as finance, healthcare, and e-commerce, where participating clients/parties often possess complementary feature sets~\citep{FL-healthcare, VFL_survey}. 
For a VFL framework with $k$ clients, each client $i$ contains a subset of features, denoted as $ \bm{X}_{i} = ( \bm{X}_{i}^{al}, \bm{X}_{i}^{nl}) $. Here, $\bm{X}_{i}^{al}$  represents aligned samples, where all clients have full local feature sets for these samples without any missing features. $\bm{X}_{i}^{nl}$ denotes non-aligned samples, where at least one client has (partially) missing features. \cref{fig:datasetting} illustrates this data partition in VFL with two clients $k=2$.

While VFL represents a promising framework by integrating disjoint local feature sets from multiple clients, its practical deployment is often hindered by two major limitations.
First, both training and inference processes in VFL depend on the complete alignment of data samples across all clients~\citep{Vfl_selfsup, Fedcvt}, i.e., only use the aligned samples $\bm{X}^{al}$. This requirement significantly limits the volume of usable data, thereby affecting both the model accuracy and the scalability of the model.
Second, the inference phase necessitates collaboration with all clients to complete the inference process, leading to significant communication overhead and making real-time or locally independent inference impractical~\citep{FedSSD, IAVFL}.
These limitations highlight an urgent need to develop a VFL framework to facilitate independent inferences and enable the efficient exploitation of non-aligned data samples (i.e., $\bm{X}^{nl}$) during both training and inference processes.

In this paper, we propose \XVFL, a novel VFL framework that exploits the non-aligned data samples with missing features and supports locally independent inference of new data samples for each client.
Particularly, we design two key modules in \XVFL: \emph{Cross Completion} (\XCom) and \emph{Decision Subspace Alignment} (\DSAlign). The \XCom module is developed to establish cross-complementary dependencies among disjoint local features contributed by different clients. Leveraging these relationships, local clients can complete their missing features through \XCom, thereby effectively increasing the volume of data available for training and inference.
The \DSAlign module is designed to align the features across all clients within the decision subspace to support locally independent inference for each client, while maintaining performance comparable to collaborative inferences involving all clients.

\begin{wrapfigure}{r}{0.5\textwidth}
	\vspace{-6mm}
	\centering
	\includegraphics[width=\linewidth]{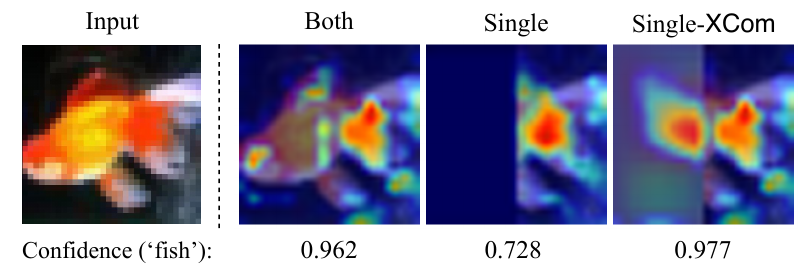}
	\vspace{-6mm}
	\caption{Effect of \XCom.}
	\label{fig:GradCAM}
\end{wrapfigure}

As illustrated in the attention heatmap (\cref{fig:GradCAM}), the image is partitioned between two clients (left and right). In the ``Single'' setting, the left-side client entirely lacks its local features. Consequently, the model relies solely on the right-side client's features resulting in a relatively low confidence score of 0.728 for the ``fish'' class.
However, with the integration of our \XCom (``Single-\XCom''), the missing features in the left-side client are completed using the information from the right-side client through \XCom. As a result, ``Single-\XCom'' significantly improves the confidence score to 0.977. Moreover, it even slightly exceeds the 0.962 confidence score obtained when using the full joint features (``Both''), showing that \XCom may also help denoise the original data and enhance inference accuracy. 
These results highlight the effectiveness of \XCom in reconstructing missing features and improving model performance.

\begin{wrapfigure}{r}{0.5\textwidth}
	\centering
	\begin{subfigure}{0.48\linewidth}
		\centering
		\includegraphics[width=\linewidth]{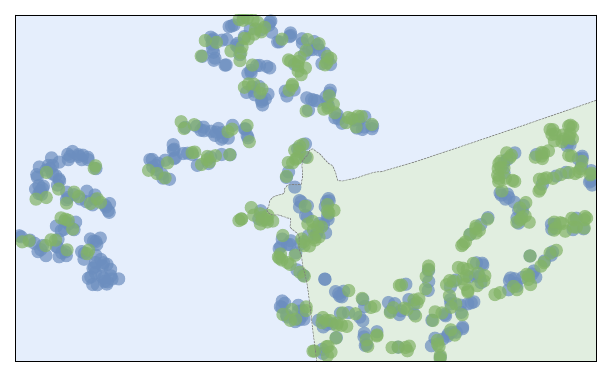}
		\caption{Vanilla}
		\label{fig:decision_vanilla}
	\end{subfigure}
	\hspace*{-3mm}
	\begin{subfigure}{0.48\linewidth}
		\centering
		\includegraphics[width=\linewidth]{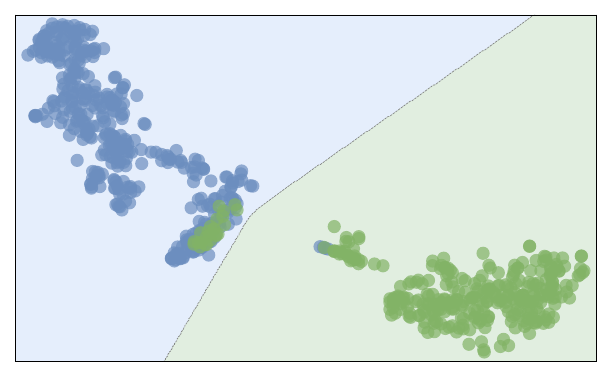}
		\caption{\DSAlign}
		\label{fig:decision_xvfl}
	\end{subfigure}	
	\vspace{-2mm}
	\caption{Effect of \DSAlign.}
	\label{fig:decision_boundry}
\end{wrapfigure}

The proposed \DSAlign module enhances \XVFL by aligning local individual features with the completed and joint features across all clients within the decision subspace. This alignment enables locally independent inference even in the presence of missing features.
As shown in~\cref{fig:decision_boundry}, the ``Vanilla'' setting (\cref{fig:decision_vanilla}) results in a poorly defined decision boundary, leading to overlapping regions and misclassifications. 
In contrast, our ``\DSAlign'' setting (\cref{fig:decision_xvfl}) yields a much clearer decision boundary, ensuring better separation between classes. This confirms the effectiveness of \DSAlign in inference under our \XVFL framework.

\subsection{Our Contributions} 
\label{subsec:Our Contributions}
We would like to highlight the following contributions:
\begin{itemize}
	
	\item We propose \XVFL, a novel VFL framework designed to handle the non-aligned data samples with (partially) missing features and to support locally independent inference for new data samples at each client. More importantly, our \XVFL introduces two key modules: Cross Completion (\XCom) and Decision Subspace Alignment (\DSAlign), which significantly enhance the ability of VFL to address more complex and practical scenarios.
	
	\item To the best of our knowledge, we are the first to introduce a practical setting with \emph{partially} missing features in VFL, where a client may retain some local features rather than fully missing all local features for non-aligned data samples. For example, consider a non-aligned data sample $X$ with $m$ features evenly partitioned between two clients, each holding $\frac{m}{2}$ features. Previous studies have only considered the fully missing setting, where one client lacks all $\frac{m}{2}$ of its local features. However, we introduce the notion \emph{missing rate}, denoted as $R_\text{miss}$, where a client may miss $R_\text{miss} \cdot \frac{m}{2}$ of its local features. In particular, the missing rate $R_\text{miss}=1$ recovers the fully missing case. Given the ubiquity of partially missing features in real-world datasets, addressing this more general and realistic setting is essential for enchancing the practical applicability of VFL. 
	
	\item Moreover, we provide theoretical convergence theorems for the algorithms used in training \XVFL, showing an $O(1/\sqrt{T})$ convergence rate for SGD-type algorithms and an $O(1/T)$ rate for PAGE-type algorithms, where $T$ denotes the number of training update steps (see Theorems~\ref{thm:sgd} and \ref{thm:page} in \cref{sec:theory}).
	
	\item Finally, we conduct extensive experiments on real-world datasets to demonstrate that \XVFL significantly outperforms existing VFL methods, e.g., achieving a 15\% improvement in accuracy on \dataset{CIFAR-10} and a 43\% improvement on the medical dataset \dataset{MIMIC-III} (see \cref{sec:experiment}). The experiments validate the practical effectiveness and superiority of \XVFL in VFL, particularly in practical scenarios involving partially missing features and locally independent inference.
	
\end{itemize}

\subsection{Related Work} \label{sec:preliminary}

Vertical Federated Learning (VFL) enables distributed learning on vertically partitioned datasets, where clients hold disjoint feature subsets of the same dataset samples~\citep{VFL_survey}.
A substantial body of work has focused on reducing communication overhead, which is critical for improving the scalability and practicality of VFL, making it a promising research direction~\citep{compressedVFL, CELU-VFL, coreset-VFL}.

In addition, several studies have explored Split Neural Network (SplitNN)-based VFL architectures to facilitate multi-client collaboration without direct data sharing~\citep{SplitNN, VFL_splitNN}. 
In these architectures, each client trains a segment of a neural network up to a predefined cut layer. The outputs at the cut layer from each client are transmitted to a central server, which holds the class labels and completes the forward and backward propagation. The resulting gradients are sent back to the clients to update their local models. This iterative process is repeated until convergence. During inference, predictions are collaboratively generated by all clients.

Research on SplitNN-based VFL mainly focused on enabling locally independent inference and handling non-aligned data samples~\citep{VFL_survey}. For instance, \citet{IAVFL} introduced a knowledge distillation framework, IAVFL, that transfers insights from joint training to local models, enabling independent inference based solely on local features. Subsequent studies extended this concept through representation-level distillation to capture inter-client feature correlations and enhance independent inference capabilities~\citep{Vfl_health, semi-VFL, Vfl_distill2}.
However, these distillation-based methods are restricted to aligned samples, thus providing limited improvement in inference performance. 

Another promising direction leverages non-aligned data to improve model performance~\citep{FedSSD, Vfl_selfsup2}. Recent efforts have applied self-supervised learning (SSL) techniques to enhance local model representations using non-aligned data samples~\citep{Vfl_selfsup, Vfl_selfsup2}. By improving the representational quality of local models, SSL typically contributes to better global model performance in VFL frameworks. However, these methods still require collaboration among all clients during the inference, restricting their practical applicability.
Besides, \citet{Fedcvt} proposed a semi-supervised learning framework, FedCVT, that estimates the corresponding embeddings/representations for the missing features and infers the pseudo labels for the missing labels, thus increasing the amount of data embeddings available for training.
\citet{LASER-VFL} proposed LASER-VFL, a VFL framework that enables training and inference with subsets of feature blocks by sharing representation models and using average aggregation with task sampling. However, it only considers the fully missing feature setting and does not support partially missing features. Moreover, it lacks the ability to complete/reconstruct missing features. 

Therefore, designing a new VFL framework that supports independent inferences and enables the efficient integration and completion of non-aligned data samples with partially missing features during both training and inference is critical for advancing the practical capabilities of VFL.

\section{\textsf{X-VFL} Framework} \label{sec:method}

In this section, we introduce \XVFL, a novel framework that overcomes the limitations of conventional VFL by supporting locally independent inference and effectively handling datasets with partially missing features.
In \cref{subsec:Framework Description}, we first describe the setup used in our \XVFL, which serves as the base for supporting both collaborative and independent inference under partially missing features.
Then, we provide an overview of the \XVFL framework by introducing its two core components: Cross Completion (\XCom) and Decision Subspace Alignment (\DSAlign).
Finally, we present the inference modes supported by our \XVFL in \cref{subsec:inference}, which include both collaborative and independent inference while effectively handling partially missing features.

\begin{figure}[t]
	\centering
	\includegraphics[width =0.83\linewidth] {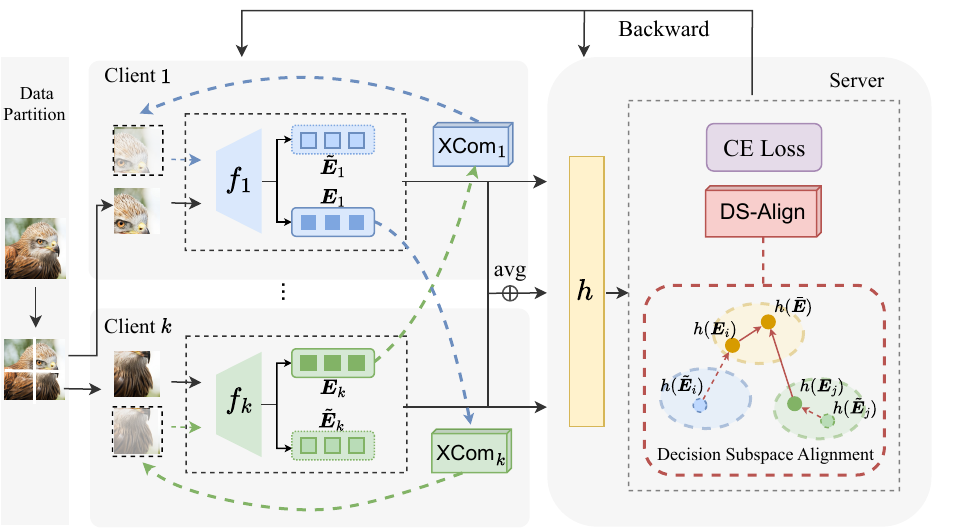}
	\caption{The framework of \XVFL.}
	\label{fig:workflow}
\end{figure}

\subsection{Framework description} \label{subsec:Framework Description}

As illustrated in \cref{fig:workflow}, our \XVFL is built upon two key modules: Cross Completion (\XCom) and Decision Subspace Alignment (\DSAlign).  Each client hosts a local/bottom model $f_{i}$ and an \XCom module. The local model $f_{i}$ processes its local features to high-level embeddings/representations, denoted as $\bm{E}_{i} = f_{i}(\bm{x}_{i})$, while the \XCom module reconstructs missing features using embeddings received from other clients. Once the embeddings, whether derived from existing features, reconstructed features, or a combination of both, are obtained, they are either directly fed into the top model $h$ or averaged before being used as input. The final loss is computed by combining the decision cross-entropy (CE) loss and the loss from the \DSAlign module, followed by standard backpropagation.

Note that previous VFL frameworks typically aggregates the embeddings via direct concatenation: $\bm{E} =[\bm{E}_{1}, \bm{E}_{2}, \cdots, \bm{E}_{k}]$, combining representations from all clients. 
However, our \XVFL performs aggregation via averaging $\bm{E}_{\text{avg}} = \mathrm{Avg}(\bm{E}_{1}, \bm{E}_{2}, \cdots, \bm{E}_{k})$, and uses  $\bm{E}_{\text{avg}}$ as the input to the top model.
This design naturally enables \emph{independent inference}, as each local embedding $E_i$ has the same dimensionality as $\bm{E}_{\text{avg}}$, allowing a single client to perform inference locally using its local model $f_i$ with the top model $h$. In contrast, under previous concatenation-based design, the global embedding $\bm{E}$ (input to the top model $h$) differs in dimensionality from individual embedding $\bm{E}_{i}$, thus requiring all clients to participate for inference (i.e., collaborative inference).

\topic{Cross Completion}
The \XCom module is designed to complete/reconstruct missing features for non-aligned data samples by leveraging information from other clients, thereby increasing the amount of usable training data and improving the inference performance.

To better illustrate this, we first consider a two-client setting with clients A and B (the general case with multiple $k$ clients is deferred to \cref{sec:general clients}).
Let $f_{a}$ and $f_{b}$ denote their bottom models, respectively. 
Their (partially) missing features can be completed using the embeddings of the other client as follows:
\begin{equation}\label{eq:vfl_xcom} 
	\tilde{\bm{X}}_{a} = \XCom_{a}(\bm{E}_{b}), \  \tilde{\bm{X}}_{b} = \XCom_{b}(\bm{E}_a),
\end{equation}
where $\bm{E}_{b}=f_{b}(\bm{X}_{b})$ and $\bm{E}_{a}=f_{a}(\bm{X}_{a})$, $\XCom_{a}$ and $\XCom_{b}$ are feature completers hosted by client A and client B, respectively. 
For the fully missing feature scenario, the completed features $\tilde{\bm{X}}_{a}$ and $\tilde{\bm{X}}_{b}$ are entirely adopted. For the partially missing feature scenario, only the components of $\tilde{\bm{X}}_{a}$ and $\tilde{\bm{X}}_{b}$ for missing features are adopted, while the remaining components should be replaced by their corresponding existing features in the original positions.
The completed features $\tilde{\bm{X}}_{a}$ and $\tilde{\bm{X}}_{b}$ are then input into their corresponding bottom models, $f_{a}$ and $f_{b}$, to generate their embeddings $\tilde{\bm{E}}_{a}$ and $\tilde{\bm{E}}_{b}$, respectively. 

Then, the classification decision loss is formulated as follows:

\begin{equation}\label{eq:l_ce_aligned_non_aligned} 
	\begin{split}
		L_{\text{decision}} = & \ell(h(\bm{E}_{a}), y) + \ell(h(\bm{E}_{b}), y)  + \ell(h(\frac{\bm{E}_{a}+\bm{E}_{b}}{2}), y) \\
		& + \ell(h(\frac{\bm{E}_{a}+\tilde{\bm{E}}_{b}}{2}), y)  + \ell(h(\frac{\tilde{\bm{E}}_{a}+\bm{E}_{b}}{2}), y),
	\end{split}
\end{equation}
where $\ell$ denotes the classification loss, e.g., cross-entropy (CE) loss. 
$\ell(h(\bm{E}_{a}), y)$ and $\ell(h(\bm{E}_{b}), y)$ ensure that the model maintains the ability to make accurate predictions based on features from a single client, thereby supporting independent inference. 
The term $\ell(h(\frac{\bm{E}_{a} + \bm{E}_{b}}{2}), y)$ reinforces the model’s capacity to leverage the combined features, enhancing collaborative inference performance.
The terms involving reconstructed embeddings, i.e., $\ell(h(\frac{\bm{E}_{a} + \tilde{\bm{E}}_{b}}{2}), y)$ and  $\ell(h(\frac{\tilde{\bm{E}}_{a} + \bm{E}_{b}}{2}), y)$, ensure that the reconstructed embeddings ($\tilde{\bm{E}}_{a}$ or $\tilde{\bm{E}}_{b}$) effectively compensate for the missing information, thus enabling robust inference in the presence of missing features.
For aligned data, all terms should be activated.
For non-aligned data, we consider two cases: 1) Client A has full local features while client B has (partially) missing features. In this case, the loss terms $\ell(h(\bm{E}_{a}), y)$ and $\ell(h(\frac{\bm{E}_{a} + \tilde{\bm{E}}_{b}}{2}), y)$ are activated; 2) The roles are reversed; client B has full local features, and client A has (partially) missing features. The activated terms are $\ell(h(\bm{E}_{b}), y)$ and $\ell(h(\frac{\tilde{\bm{E}}_{a} + \bm{E}_{b}}{2}), y)$. 

By explicitly leveraging reconstructed features, \XCom effectively addresses the challenges of training and inference with (partially) missing features in VFL.

\topic{Decision Subspace Alignment}
\DSAlign unifies the alignment between reconstructed and existing features for aligned data, as well as between local individual features and the joint averaged features aggregated from all clients, within the decision subspace. This reinforces the effectiveness of \XCom and further enhances the performance of independent inference.

To enforce consistency between reconstructed features and their corresponding existing features, ensuring accurate and reliable feature completion, \DSAlign introduces the following alignment loss within the decision subspace:
\begin{equation}\label{eq:DSAlign_1}
	L_{\text{DSAlign}_{1}} = \ell(h(\tilde{\bm{E}}_{a}), h(\bm{E}_{a}) ) + \ell(h(\tilde{\bm{E}}_{b}), h(\bm{E}_{b})),
\end{equation}
where $\ell$ represents a similarity loss function, e.g., mean square error (MSE), which ensures that inference based on reconstructed embeddings aligns closely with the existing embeddings.

To enhance the performance of independent inference, \DSAlign aligns local individual features and joint averaged features from all clients within the decision subspace. This alignment enables more accurate independent inference and is formulated as:

\begin{equation}\label{eq:DSAlign_2}
	L_{\text{DSAlign}_{2}} = \ell( h(\bm{E}_{a}), h(\frac{\bm{E}_{a}+\bm{E}_{b}}{2})) + \ell(h(\bm{E}_{b}), h(\frac{\bm{E}_{a}+\bm{E}_{b}}{2})).
\end{equation}

This dual-component approach reinforces both feature completion and independent inference, thus enhancing the overall performance and scalability of \XVFL.

Finally, by integrating \XCom and \DSAlign, we formulate the overall loss function for our \XVFL framework as follows:
\begin{equation}\label{eq:overall}
	L = L_{\text{decision}} + \lambda_1 L_{\text{DSAlign}_{1}} + \lambda_2 L_{\text{DSAlign}_{2}}.
\end{equation}
Here, $L_{\text{decision}}$ (see \cref{eq:l_ce_aligned_non_aligned}) denotes the classification decision loss for both aligned and non-aligned data samples, while $L_{\text{DSAlign}_{1}}$ and $L_{\text{DSAlign}_{2}}$  (see \cref{eq:DSAlign_1} and \cref{eq:DSAlign_2}) are the two loss components introduced by our \DSAlign, weighted by the hyperparameters $\lambda_1$ and $\lambda_2$, respectively.

\subsection{Inference modes of \textsf{X-VFL}}
\label{subsec:inference}

\begin{figure}[b]
	\centering
	\hspace{-5mm} 
	\begin{subfigure}{0.33\textwidth}
		\centering
		\includegraphics[width=\textwidth]{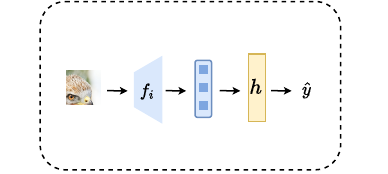} 
		\caption{Independent inference}
		\label{fig:Prediction_alone} 
	\end{subfigure}
	\hspace{-4mm}
	\begin{subfigure}{0.33\textwidth}
		\centering
		\includegraphics[width=\textwidth]{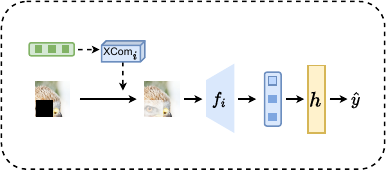} 
		\caption{Ind. infer. with missing features}
		\label{fig:Prediction_alone_missing} 
	\end{subfigure}
	\hspace{1mm}
	\begin{subfigure}{0.33\textwidth}
		\centering
		\includegraphics[width=\textwidth]{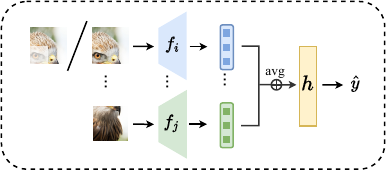} 
		\caption{Collaborative inference}
		\label{fig:Prediction_collaborateive} 
	\end{subfigure}
	\caption{Different inference modes of \XVFL for independent and collaborative inference, with/without missing features.}
	\label{fig:Prediction_mode}
\end{figure}

As illustrated in \cref{fig:Prediction_mode}, the \XVFL framework supports multiple inference modes to highlight the flexibility and scalability of \XVFL:
\begin{enumerate}
	\item \textbf{Independent inference without missing features} (\cref{fig:Prediction_alone}): 
	Each client independently generates predictions using only its local features and model, without requiring cross-client communication during inference.
	
	\item \textbf{Independent inference with missing features} (\cref{fig:Prediction_alone_missing}):  
	Each client leverages its \XCom module to reconstruct missing features and produce predictions independently.
	
	\item \textbf{Collaborative inference with/without missing features} (\cref{fig:Prediction_collaborateive}):
	All clients contribute their feature embeddings (obtained either from the original data or from the reconstructed data via \XCom) to the central server, enabling comprehensive and robust predictions through collaborative inference.
\end{enumerate}

\section{Theoretical Results} \label{sec:theory}

In this section, we provide theoretical convergence theorems for our \XVFL framework which is formulated as the following optimization problem (see \cref{eq:overall} in \cref{subsec:Framework Description}):
\begin{align}
	\min_{\bm{\theta}} L(\bm{\theta}) := L_{\text{decision}} + \lambda_1 L_{\text{DSAlign}_{1}} + \lambda_2 L_{\text{DSAlign}_{2}},
\end{align}
where $\bm{\theta} \in \R^d$ denotes the \XVFL model parameters, $\lambda_1 \in \R$ and $\lambda_2 \in \R$ are two hyperparameters.
To present the thorems, we first introduce the necessary notation and assumptions. 

\subsection{Notation and assumptions}\label{subsec:assumption}

Let $\Delta_0:=L(\bm{\theta}^0)-L^*$, where $L^*:=\min_{\bm{\theta}} L(\bm{\theta})$. 
Let $\nabla L(\bm{\theta})$ denote the gradient of function $L$ at point $\bm{\theta}$, $\tilde{\nabla} L(\bm{\theta})$ and $\tilde{\nabla}_b L(\bm{\theta})$ denote its stochastic gradient and minibatch stochastic gradients with size $b$. 
Let $[n]$ denote the set $\{1,2,\cdots,n\}$ and $\n{\cdot}$ denote the Euclidean norm for a vector.
Let $\inner{\bm{u}}{\bm{v}}$ denote the inner product of two vectors $\bm{u}$ and $\bm{v}$. 
We use $O(\cdot)$ to hide absolute constants. 

In order to prove the convergence results, one usually needs the following standard smoothness assumption and stochastic gradient variance assumption (see e.g., \citealp{ghadimi2016mini, fang2018spider, PAGE, li2020unified}).
\begin{assumption}[Average smoothness]\label{asp:avgsmooth}
	A function $f:\R^d\to \R$ is average $\beta$-smooth if 
	\begin{equation}\label{eq:avgsmooth}
		\E[\ns{\tilde{\nabla} L(\bm{\theta}_1) - \tilde{\nabla} L(\bm{\theta}_2)}]\leq \beta^2 \ns{\bm{\theta}_1-\bm{\theta}_2}, \quad \forall \bm{\theta}_1, \bm{\theta}_2 \in \R^d.
	\end{equation}
\end{assumption}

\begin{assumption}[Bounded variance]\label{asp:bv}
	The stochastic gradient $\tilde{\nabla} L(\bm{\theta})$ is unbaised and has bounded variance 
	\begin{equation}\label{eq:bv}
		\E[\ns{\tilde{\nabla} L(\bm{\theta})-{\nabla} L(\bm{\theta})}]\leq \sigma^2, \quad \forall \bm{\theta} \in \R^d.
	\end{equation}
\end{assumption}

\subsection{Convergence results}\label{subsec:convergence}
Now we present the convergence theorems for training \XVFL to find a suitable target parameter $\hat{\bm{\theta}}$. The convergence theorems can also indicate the communication complexity and computation complexity of different algorithms (such as standard stochastic gradient descent (SGD) and the optimal variance-reduced PAGE~\citep{PAGE}) for training \XVFL.

\begin{theorem}[Convergence for SGD-type algorithms]\label{thm:sgd}
	Suppose that Assumptions \ref{asp:avgsmooth} and \ref{asp:bv} hold. For SGD-type algorithms, e.g., update $\bm{\theta}^{t+1} = \bm{\theta}^t - \eta \tilde{\nabla} L(\bm{\theta}^t)$, after $T$ steps, we have
		$	\frac{1}{T}\sum_{t=0}^{T-1} \E\ns{\nabla L(\bm{\theta}^t)} \\ 
		\leq \frac{\Delta_0}{(\eta - \beta\eta^2/2)T } + \frac{\beta\eta^2\sigma^2}{2\eta - \beta\eta^2}$.
	Choose learning rate $\eta \leq \min\{\frac{2}{\beta}, \sqrt{\frac{2\Delta_0}{\beta\sigma^2T}}\}$, we obtain 
	\begin{align}\label{eq:sgd2}
		\frac{1}{T}\sum_{t=0}^{T-1} \E\ns{\nabla L(\bm{\theta}^t)} \leq O\left(\frac{1}{\sqrt{T}}\right).
	\end{align}
	This means that after $T=O(\frac{1}{\epsilon^4})$ steps, SGD-type algorithms can find a suitable parameter $\hat{\bm{\theta}}$ for \XVFL such that $\E\ns{\nabla L(\hat{\bm{\theta}})} \leq \epsilon^2$, where $\epsilon$ denotes the convergence error.
\end{theorem}

Note that each update step requires one round of communication in \XVFL, so the total number of communication rounds is $T=O(\frac{1}{\epsilon^4})$. Also, each step incurs a computational cost corresponding to a stochastic gradient computation $\tilde{\nabla} L(\bm{\theta}^t)$.

We note that the large number of communication rounds is mainly due to the variance in stochastic gradients. 
Thus we also show that the optimal variance-reduced PAGE method~\citep{PAGE} can largely reduce the number of communication rounds, i.e., from $T=O(\frac{1}{\epsilon^4})$ to $T=O(\frac{1}{\epsilon^2})$ by a factor of $\frac{1}{\epsilon^2}$.
A simplified PAGE update at step $t$ is 
\begin{align}\label{eq:pageupdate}
	 \bm{\theta}^{t+1} = \bm{\theta}^t - \eta \bm{g}^t, ~~\text{where}~
	 \bm{g}^{t} = \begin{cases}
		 		\tilde{\nabla}_b L(\bm{\theta}^{t}) &\text{with probability } p\\
		 		\bm{g}^{t-1}+ \tilde{\nabla}_{b'} L(\bm{\theta}^{t})- \tilde{\nabla}_{b'} L(\bm{\theta}^{t-1}) &\text{with probability } 1-p
		 	\end{cases}.  
\end{align}
It uses minibatch SGD update $\tilde{\nabla}_b L(\bm{\theta}^{t})$ with probability $p$, and reuses the previous gradient $\bm{g}^{t-1}$ with a small adjustment (lower computation cost if $b'\ll b$). This update reduces the variance in stochatic gradients during training and thus significantly decreases the total number of  communication rounds, i.e., by a fator of $\frac{1}{\epsilon^2}$.

\begin{theorem}[Convergence for PAGE-type algorithms]\label{thm:page}
	Suppose that Assumptions \ref{asp:avgsmooth} and \ref{asp:bv} hold. For PAGE-type algorithms, e.g., update as \cref{eq:pageupdate}, after $T$ steps, we have
		$\frac{1}{T}\sum_{t=0}^{T-1} \E\ns{\nabla L(\bm{\theta}^t)} \\ 
		\leq \frac{2\Delta_0}{\eta T } + \frac{\sigma^2}{pbT} + \frac{\sigma^2}{b}$.
	Choose learning rate $\eta \leq \frac{1}{2\beta}$, minibatch sizes $b=\frac{2\sigma^2}{\epsilon^2}$, $b' \leq \sqrt{b}$ and probability $p=\frac{b'}{b+b'}$, we obtain 
	\begin{align}\label{eq:page2}
		\frac{1}{T}\sum_{t=0}^{T-1} \E\ns{\nabla L(\bm{\theta}^t)} \leq O\left(\frac{1}{T}\right).
	\end{align}
	This means that after $T=O(\frac{1}{\epsilon^2})$ steps, PAGE-type algorithms can find a suitable parameter $\hat{\bm{\theta}}$ for \XVFL such that $\E\ns{\nabla L(\hat{\bm{\theta}})} \leq \epsilon^2$, where $\epsilon$ denotes the convergence error.
\end{theorem}

Similarly, each update step requires one round of communication in \XVFL, so the total number of communication rounds is $T=O(\frac{1}{\epsilon^2})$. Also, each step incurs a computational cost corresponding to a minibatch stochastic gradients computation with expected size $pb+(1-p)b' = \frac{b'b}{b+b'}+\frac{bb'}{b+b'} \leq 2b'$.

\section{Experiments} \label{sec:experiment}

In this section, we present a comprehensive empirical evaluation of the proposed \XVFL framework using 6 real-world datasets (three image datasets: \dataset{CIFAR-10}~\citep{CIFAR10}, \dataset{TinyImageNet} (restricted to five classes)~\citep{ImageNet}, \dataset{UTKFace}~\citep{utkface}, and three tabular datasets: \dataset{MIMIC-III}~\citep{MIMIC}, \dataset{Bank}~\citep{bank}, and \dataset{Avazu}~\citep{Fedcvt}).
We compare our \XVFL against four methods (Vanilla Standalone, Vanilla VFL, IAVFL~\citep{IAVFL}, FedCVT~\citep{Fedcvt}).
We conduct experiments under a variety of practical scenarios, including: varying feature missing rates within each client (see~\cref{subsec:Results_missing features}), varying overlap ratios of aligned data samples (see~\cref{subsec:Results_overlap}), and imbalanced training data across clients (see~\cref{subsec:Results_unbalance}).
More details of the experimental setup are provided in \cref{sec:experiment details}.
Additional experiments on more datasets and in the multiple clients setting are deferred to \cref{sec:2 Clients Supplementary Details} and \ref{sec:general clients}, respectively.

\begin{figure}[!t]
	\centering
	\includegraphics[width = 1.0\linewidth]{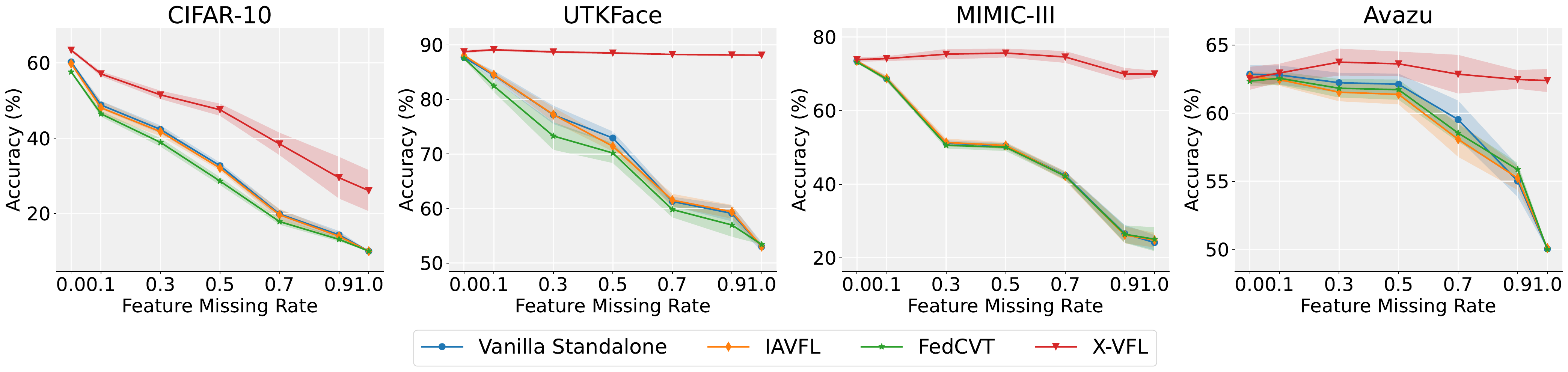}
	\caption{Performance comparison under varying feature missing rates in the independent inference mode on the \dataset{CIFAR-10}, \dataset{UTKFace}, \dataset{MIMIC-III}, and \dataset{Avazu} datasets.}
	\label{fig:performance_ind_missing_features}
	%
		\vspace{5mm}
	\centering
	\includegraphics[width = 1.0\linewidth] {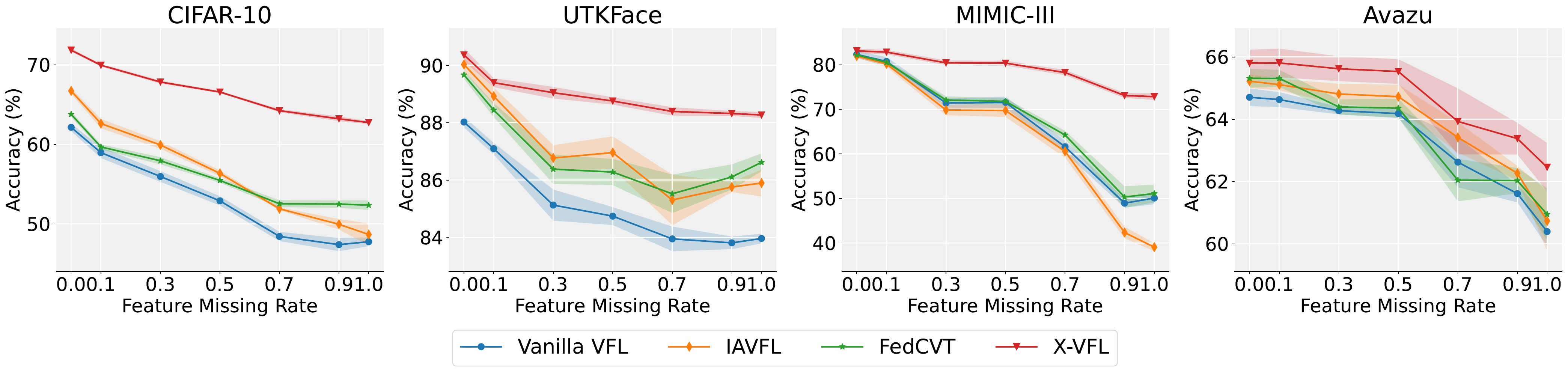}
	\caption{Performance comparison under varying feature missing rates in the collaborative inference mode on the \dataset{CIFAR-10}, \dataset{UTKFace}, \dataset{MIMIC-III}, and \dataset{Avazu} datasets.}
	\label{fig:performance_col_missing_features}
%
		\vspace{5mm}
	\centering
	\includegraphics[width = 1.0\linewidth]{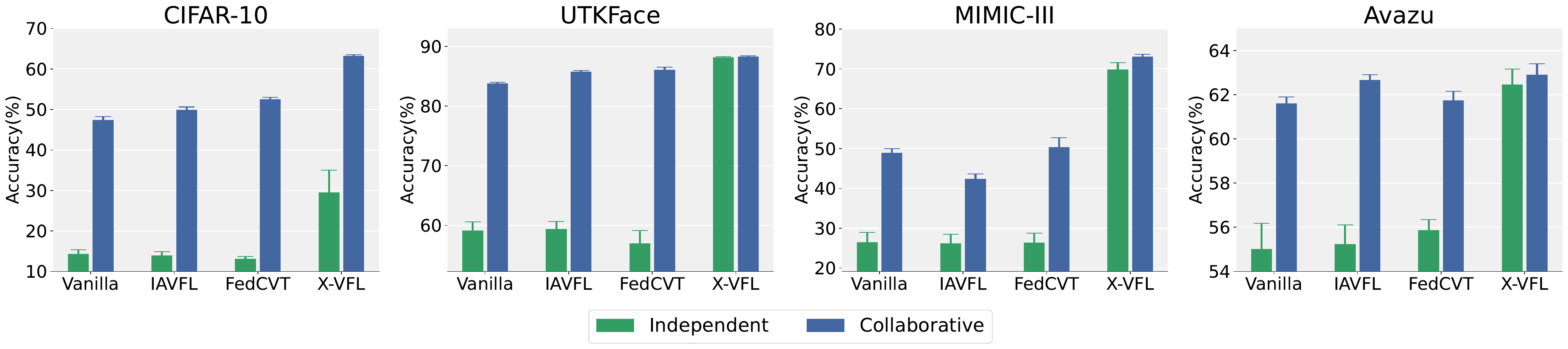}
	\caption{Performance comparison between independent and collaborative inference modes under feature missing rate $R_\text{miss}=0.9$ on the \dataset{CIFAR-10}, \dataset{UTKFace}, \dataset{MIMIC-III}, and \dataset{Avazu} datasets.}
	\label{fig:performance_diff}
\end{figure}

\subsection{Results under varying feature missing rates}
\label{subsec:Results_missing features}

The experimental results on \dataset{CIFAR-10}, \dataset{UTKFace}, \dataset{MIMIC-III}, and \dataset{Avazu} are summarized in \cref{fig:performance_ind_missing_features} and \cref{fig:performance_col_missing_features}, corresponding to the independent and collaborative inference modes, respectively. 
A comparison of performance between these two modes at a missing rate $R_\text{miss}=0.9$  is presented in \cref{fig:performance_diff}.

As shown in \cref{fig:performance_ind_missing_features} and \cref{fig:performance_col_missing_features}, \XVFL consistently outperforms all methods across all datasets and feature missing rates, in both independent and collaborative modes. 
Moreover, the performance gap further widens as the missing rate increases. For instance, on the \dataset{UTKFace} dataset (in \cref{fig:performance_ind_missing_features}) with a feature missing rate $R_\text{miss}=0.9$, the independent prediction accuracies of the three baseline methods drop from nearly 90\% (at $R_\text{miss}=0$, i.e., no missing data) to around 60\%, indicating a performance degradation of 30\%. In contrast, \XVFL maintains high accuracy under the same condition, with a decrease of less than 0.5\% compared to its own performance at $R_\text{miss}=0$.

\cref{fig:performance_diff} highlights the advantage of \XVFL over the baseline methods by demonstrating its ability to effectively narrow the accuracy gap between independent and collaborative inference modes.
For instance, on the \dataset{UTKFace} dataset, \XVFL exhibits a tiny performance gap of less than 0.2\% between independent inference (88.15\%) and collaborative inference (88.32\%). In contrast, baseline methods suffer a significant accuracy drop of over 20\% when switching from collaborative to independent inference.
A similar trend is observed across other datasets, indicating \XVFL's robustness in maintaining consistent performance.

\subsection{Results under varying overlap ratios of aligned data samples}
\label{subsec:Results_overlap}

\begin{figure}[t]
	\centering
	\begin{subfigure}{0.48\textwidth}
		\centering
		\includegraphics[width=\linewidth]{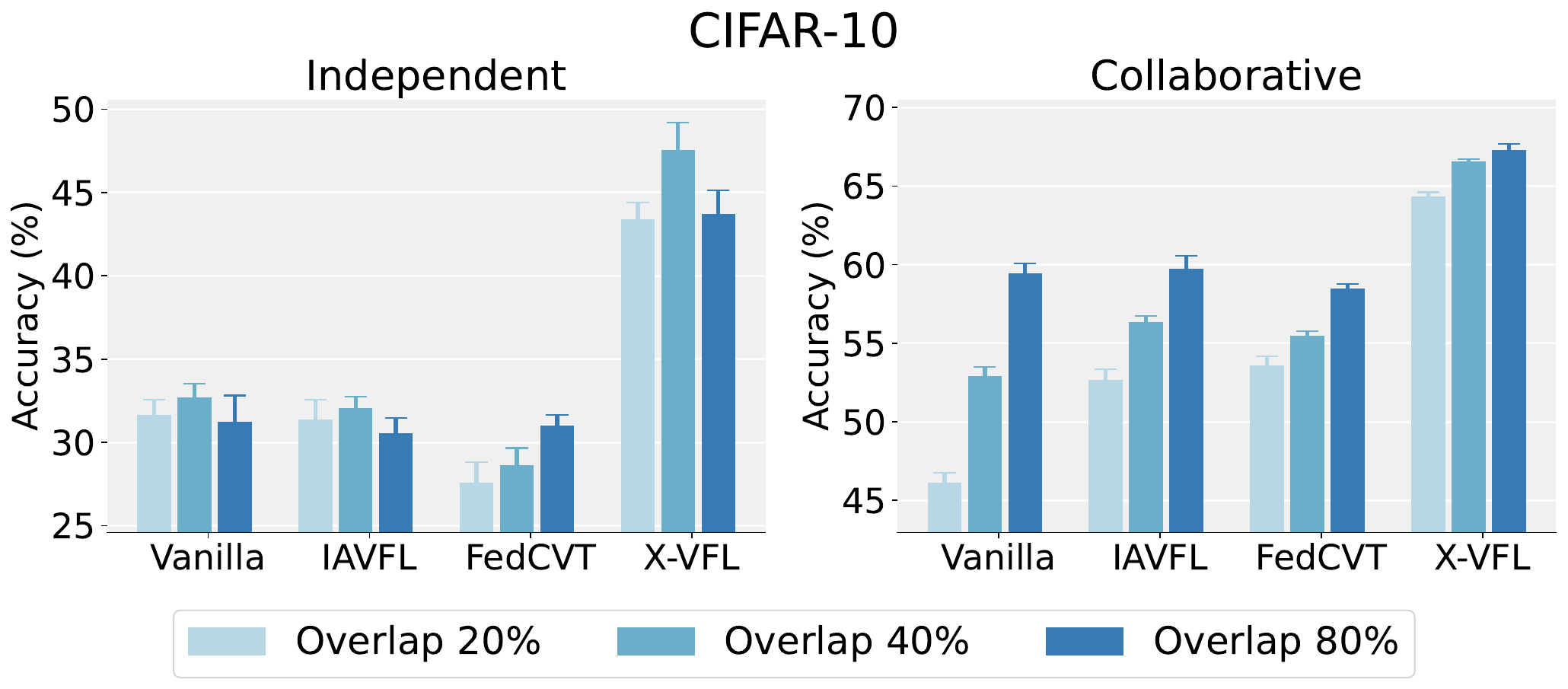}
		\caption{\dataset{CIFAR-10} dataset}
		\label{fig:difOverlap_cifar10}
	\end{subfigure}\hspace{4mm}
	\begin{subfigure}{0.48\textwidth}
		\centering
		\includegraphics[width=\linewidth]{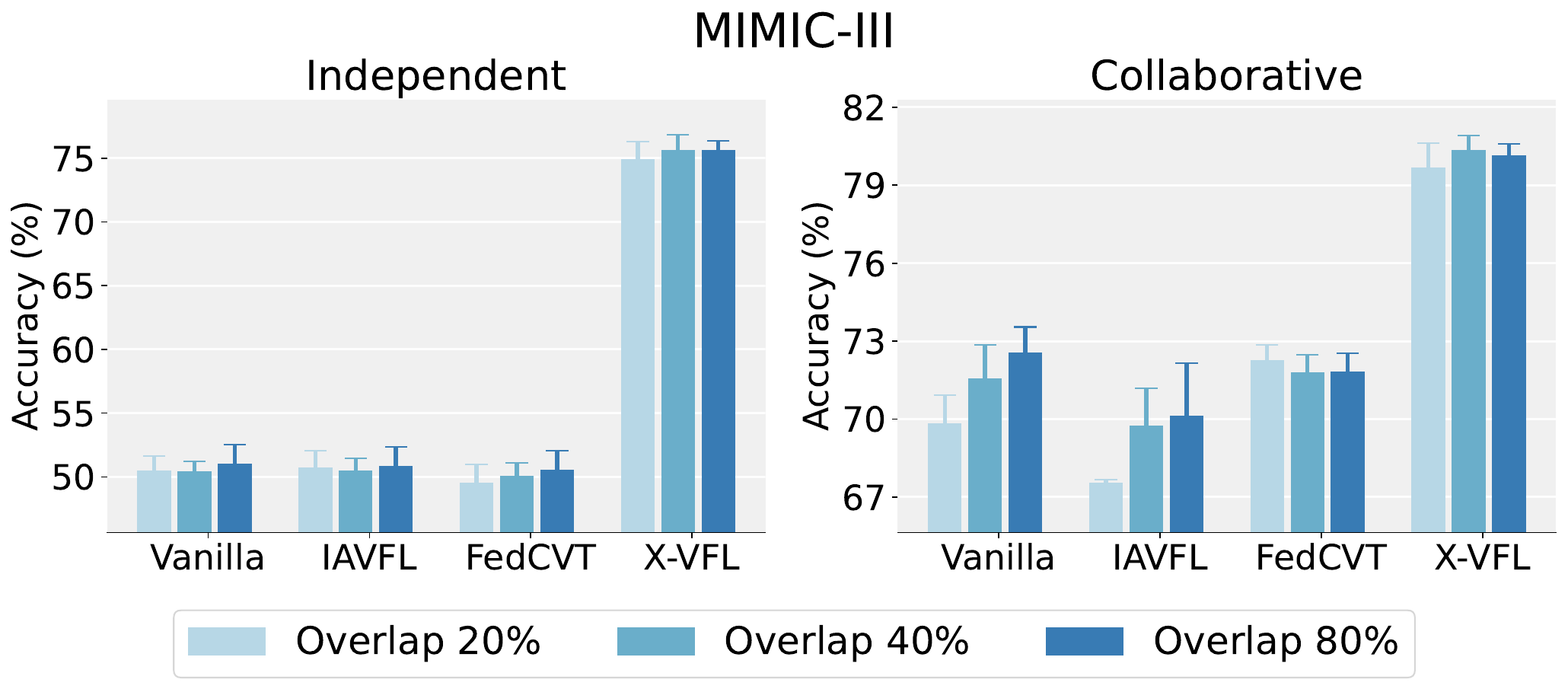}
		\caption{\dataset{MIMIC-III} dataset}
		\label{fig:difOverlap_mimic}
	\end{subfigure}
	\caption{Performance comparison under varying overlap ratios on the \dataset{CIFAR-10} and \dataset{MIMIC-III} datasets.}
	\label{fig:bar_chart_2_cifar10_mimic}
\end{figure}

To evaluate the effect of varying overlap ratios of aligned data samples, we test these methods with 20\%, 40\%, and 80\% aligned samples, under a fixed feature missing rate of $R_\text{miss}=0.5$.
The performance results on \dataset{CIFAR-10} and \dataset{MIMIC-III} are presented in \cref{fig:difOverlap_cifar10} and \cref{fig:difOverlap_mimic}, respectively. 

As illustrated, \XVFL consistently outperforms the baseline methods across all overlap ratios in both independent and collaborative inference modes. 
For instance, on the \dataset{CIFAR-10} dataset (\cref{fig:difOverlap_cifar10}), 
\XVFL achieves over 45\% accuracy in the independent inference mode at a 40\% overlap ratio, whereas the three baseline methods all fall below 35\%. Similarly, in the collaborative inference mode, \XVFL attains over 65\% accuracy at the same 40\% overlap ratio, compared to around 55\% for the baseline methods.

Also, a similar trend is observed on the \dataset{MIMIC-III} dataset (\cref{fig:difOverlap_mimic}), further demonstrating \XVFL's robustness and superior performance, particularly under low overlap ratios.

\subsection{Results under imbalanced training data across clients}
\label{subsec:Results_unbalance}

\begin{wrapfigure}{r}{0.5\textwidth}
	\centering
	\includegraphics[width=\linewidth]{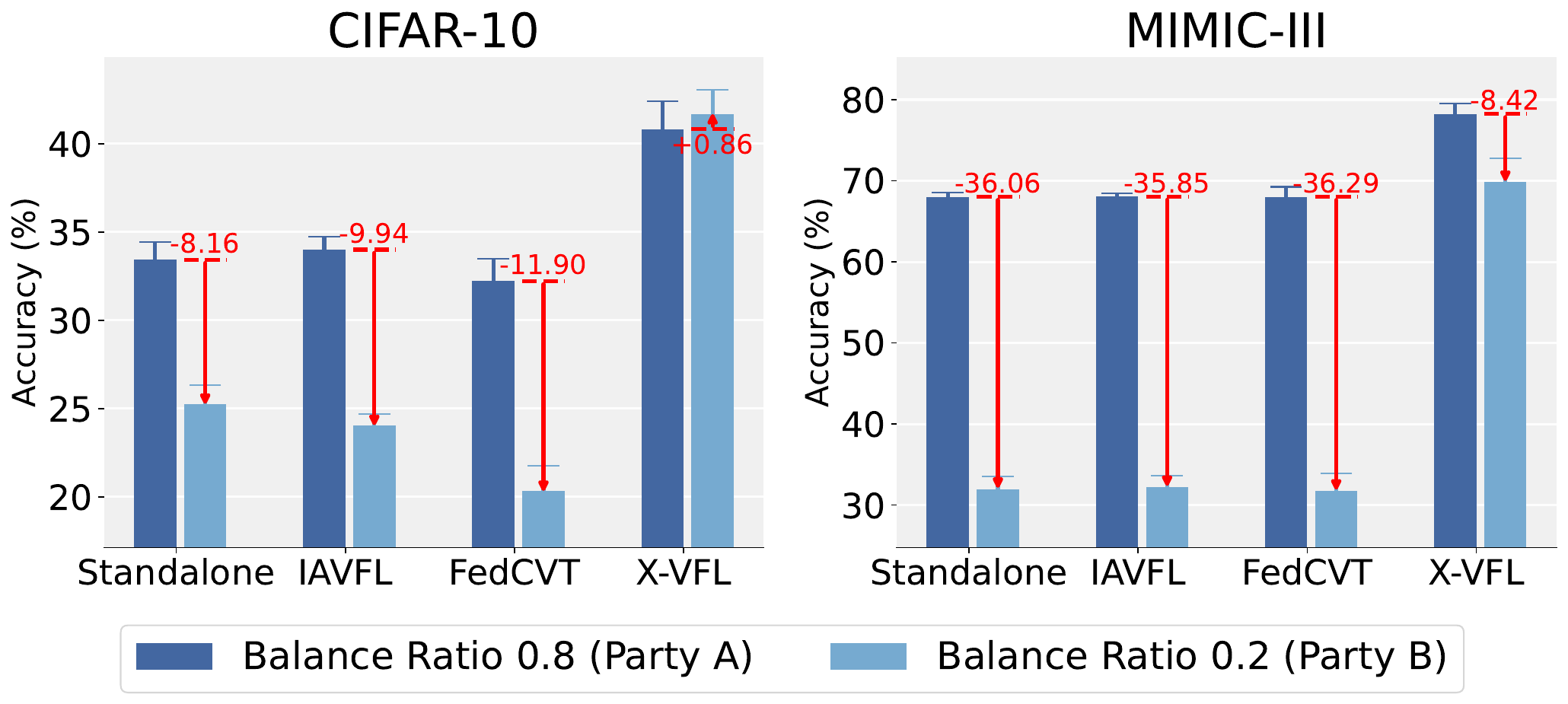}
	\vspace{-5mm}
	\caption{Performance comparison under data imbalance on the \dataset{CIFAR-10} and \dataset{MIMIC-III} datasets.}
	\label{fig:balance_insensitive}
\end{wrapfigure}

To evaluate the impact of data imbalance between clients for these methods,
we consider a 20\% data imbalance setting, where client A and client B hold 80\% and 20\% of the total data, respectively, including both aligned and non-aligned samples. 
This setup enables a thorough evaluation of the impact of imbalanced training data on classification performance. 
The performance gap between the data-rich and data-poor clients also offers an interesting point of analysis. 
Experimental results for independent inference under a feature missing rate of $R_\text{miss}=0.5$ are presented in \cref{fig:balance_insensitive}.

As demonstrated in \cref{fig:balance_insensitive}, the baseline methods exhibit substantial performance gaps between the two clients, particularly on the \dataset{MIMIC-III} dataset, where the data-rich client A consistently outperforms the data-poor client B by 36.06\%, 35.85\%, and 36.29\%. 
These gaps highlight the inability of baseline methods to effectively handle data imbalance, leading to significantly degraded performance for the data-poor client B. 
In contrast, our \XVFL substantially reduces this gap to just 8.42\%, demonstrating its superior ability to balance the performance between data-rich and data-poor clients, by completing the missing features through the \XCom module. 
A similar yet more compelling trend is observed on the \dataset{CIFAR-10} dataset. While baseline methods fail to adequately support the data-poor client B, \XVFL not only narrows the performance gap but also slightly improves the accuracy of the data-poor client B. 
This unexpected gain can be attributed to the design of the \XCom module, which leverages the richer data from client A to enhance the feature representations for the data-poor client B by reconstructing its missing features, thus improving classification performance despite the severe data imbalance. 


\section{Conclusion}
\label{sec:conclusion}

In this paper, we proposed \XVFL, a novel VFL framework that addresses key challenges of conventional VFLs by effectively handling datasets with partially missing features and enabling locally independent inference at each client. 
In particular, \XVFL introduces two key modules: Cross Completion (\XCom) and Decision Subspace Alignment (\DSAlign). 
\XCom is designed to complete missing features for non-aligned data samples by exploiting cross-client information, thereby effectively increasing the volume of data available for training and inference.
\DSAlign aligns local features with completed and global features across all clients within the decision subspace, enabling each client to perform locally independent inference, even in the presence of missing features.
In addition, theoretical convergence theorems are estabilished for different algorithms used in training \XVFL.
Extensive experiments on real-world datasets demonstrate that \XVFL significantly outperforms existing VFL methods, validating its practical effectiveness and superiority in addressing key challenges such as missing features, locally independent inference, and data imbalance.

\section*{Acknowledgments}
This research was supported by the Singapore Ministry of Education (MOE) Academic Research Fund (AcRF) Tier 1 grant.

\bibliographystyle{plainnat}
\bibliography{Reference}

\newpage
\appendix
\section{Missing Proofs}~\label{sec:missing proofs}
In this appendix, we provide the detailed proofs for our convergence Theorems~\ref{thm:sgd} and \ref{thm:page} in \cref{sec:proofofsgd} and \cref{sec:proofofpage}, respectively.

\subsection{Proof of \cref{thm:sgd}}~\label{sec:proofofsgd}
According to SGD-type update $\bm{\theta}^{t+1} = \bm{\theta}^t - \eta \tilde{\nabla} L(\bm{\theta}^t)$ and the smoothness \cref{asp:avgsmooth}, we have
\begin{align}
	\E_t[L(\bm{\theta}^{t+1})] 
	&\leq \E_t[L(\bm{\theta}^t) + \inner{\nabla L(\bm{\theta}^t)}{\bm{\theta}^{t+1} - \bm{\theta}^t} + \frac{\beta}{2}\ns{\bm{\theta}^{t+1}-\bm{\theta}^t}]\\
	&= \E_t[L(\bm{\theta}^t) -\eta \inner{\nabla L(\bm{\theta}^t)}{\tilde{\nabla} L(\bm{\theta}^t)} + \frac{\beta\eta^2}{2}\ns{\tilde{\nabla} L(\bm{\theta}^t)}] \\
	&\leq L(\bm{\theta}^t) - (\eta -\frac{\beta\eta^2}{2})\ns{\nabla L(\bm{\theta}^t)} + \frac{\beta\eta^2\sigma^2}{2}, \label{eq:usebv} 
\end{align}
where $\E_t$ takes the expectation conditioned on all history before step $t$, and \cref{eq:usebv} uses \cref{asp:bv}.

Summing up \cref{eq:usebv} from $t=0$ to $T-1$ and rearranging terms, we get
\begin{align}
	(\eta -\frac{\beta\eta^2}{2})\sum_{t=0}^{T-1}\E\ns{\nabla L(\bm{\theta}^t)} 
	&\leq L(\bm{\theta}^0) - \E[L(\bm{\theta}^t)] + \frac{\beta\eta^2\sigma^2 T}{2} \\
	\frac{1}{T}\sum_{t=0}^{T-1}\E\ns{\nabla L(\bm{\theta}^t)} 
	&\leq \frac{\Delta_0}{(\eta-\beta\eta^2/2)T} + \frac{\beta\eta^2\sigma^2}{2\eta - \beta\eta^2}, \label{eq:usedelta0}
\end{align}
where $\Delta_0:=L(\bm{\theta}^0) -L^*$.
Note that 
$\frac{\beta\eta^2\sigma^2}{2\eta - \beta\eta^2} \leq \frac{\Delta_0}{(\eta-\beta\eta^2/2)T}$
if $\eta \leq \sqrt{\frac{2\Delta_0}{\beta\sigma^2T}}$.
Thus it is not hard to obtain, by choosing the learning rate $\eta \leq \min\{\frac{2}{\beta}, \sqrt{\frac{2\Delta_0}{\beta\sigma^2T}}\}$,
\begin{align}
	\frac{1}{T}\sum_{t=0}^{T-1}\E\ns{\nabla L(\bm{\theta}^t)} 
	&\leq O\left(\sqrt{\frac{\beta\Delta_0\sigma^2}{T}} \right)
	= O\left(\frac{1}{\sqrt{T}}\right) 
	= \epsilon^2.
\end{align}
This means that after $T=O(\frac{1}{\epsilon^4})$ steps, SGD-type algorithms can find a suitable parameter $\hat{\bm{\theta}}$ for \XVFL such that $\E\ns{\nabla L(\hat{\bm{\theta}})} \leq \epsilon^2$, where $\epsilon$ denotes the convergence error.
\qedb

\subsection{Proof of \cref{thm:page}}\label{sec:proofofpage}
The variance in stochastic gradients leads to a large number of communication rounds (i.e., update steps $T$). We now prove that the optimal variance-reduced PAGE-type method~\citep{PAGE} effectively reduces the variance and thus largely decreases the total number of communication rounds (\cref{thm:page}).

According to simplified PAGE update step \cref{eq:pageupdate} and  Lemma 4 of PAGE~\citep{PAGE}, we have the following variance reduction lemma:
\begin{lemma}
	Suppose that Assumptions~\ref{asp:avgsmooth} and \ref{asp:bv} hold. For the probabilistic gradient estimator $\bm{g}^t$ defined in \cref{eq:pageupdate}, we have 
	\begin{align}
		\E_t[\ns{\bm{g}^t - \nabla L(\bm{\theta}^{t})}] 
		\leq (1-p)\ns{\bm{g}^{t-1} - \nabla L(\bm{\theta}^{t-1})} 
		+ \frac{(1-p)\beta^2}{b'}\ns{\bm{\theta}^{t}-\bm{\theta}^{t-1}} 
		+ \frac{p\sigma^2}{b}. \label{eq:variance-reduce}
	\end{align}
\end{lemma}
This lemma indicates that the variance is roughly reduced by a factor $1-p$ after each update step.
Then according to the descent Lemma 2 of PAGE~\citep{PAGE}, we have 
\begin{align}
	L(\bm{\theta}^{t+1}) \leq L(\bm{\theta}^t) - \frac{\eta}{2} \ns{\nabla f(\bm{\theta}^t)} 
	- (\frac{1}{2\eta} - \frac{\beta}{2})\ns{\bm{\theta}^{t+1} - \bm{\theta}^t} 
	+ \frac{\eta}{2}\ns{\bm{g}^t-\nabla L(\bm{\theta}^t)}. \label{eq:descent}
\end{align}

Taking the expectation and combining \cref{eq:descent} with multiple  \cref{eq:variance-reduce}, we obtain 
\begin{align}
	&\E[L(\bm{\theta}^{t+1}) - L^* + \frac{\eta}{2p}\ns{\bm{g}^{t+1} - \nabla L(\bm{\theta}^{t+1})}] \notag\\
	&\leq \E\Big[L(\bm{\theta}^{t}) - L^* 
	- \frac{\eta}{2}\ns{\nabla L(\bm{\theta}^{t})} 
	- (\frac{1}{2\eta} - \frac{\beta}{2})\ns{\bm{\theta}^{t+1} - \bm{\theta}^t} 
	+ \frac{\eta}{2}\ns{\bm{g}^t-\nabla L(\bm{\theta}^t)} \newll
	+\frac{\eta}{2p}\Big((1-p)\ns{\bm{g}^{t} - \nabla L(\bm{\theta}^{t})} 
	+ \frac{(1-p)\beta^2}{b'}\ns{\bm{\theta}^{t+1}-\bm{\theta}^{t}} 
	+ \frac{p\sigma^2}{b}\Big)\Big] \\
	&=\E\Big[ L(\bm{\theta}^{t}) - L^*  + \frac{\eta}{2p}\ns{\bm{g}^{t} - \nabla L(\bm{\theta}^{t})}
	-\frac{\eta}{2}\ns{\nabla L(\bm{\theta}^{t})} 
	+ \frac{\eta\sigma^2}{2b} \newll
	-(\frac{1}{2\eta} - \frac{\beta}{2} - \frac{(1-p)\eta\beta^2}{2pb'})\ns{\bm{\theta}^{t+1} - \bm{\theta}^t} 
	\Big] \\
	&\leq \E\Big[ L(\bm{\theta}^{t}) - L^*  + \frac{\eta}{2p}\ns{\bm{g}^{t} - \nabla L(\bm{\theta}^{t})}
	-\frac{\eta}{2}\ns{\nabla L(\bm{\theta}^{t})} 
	+ \frac{\eta\sigma^2}{2b}\Big], \label{eq:use-eta1}
\end{align}
where \cref{eq:use-eta1} holds due to $\frac{1}{2\eta} - \frac{\beta}{2} - \frac{(1-p)\eta\beta^2}{2pb'}\geq 0$ by choosing learning rate $\eta\leq\frac{1}{\beta(1+\sqrt{(1-p)/(pb')})}$.

Summing up \cref{eq:use-eta1} from $t=0$ to $T-1$ and rearranging terms, we get
\begin{align}
	\frac{\eta}{2}\sum_{t=0}^{T-1}\E\ns{\nabla L(\bm{\theta}^t)} 
	&\leq \E\Big[L(\bm{\theta}^0) - L^* + \frac{\eta}{2p}\ns{\bm{g}^0-\nabla L(\bm{\theta}^0)} 
	+\frac{\eta\sigma^2 T}{2b}\Big] \\
	\frac{1}{T}\sum_{t=0}^{T-1}\E\ns{\nabla L(\bm{\theta}^t)} 
	&\leq \frac{2\Delta_0}{\eta T} + \frac{\sigma^2}{pbT} + \frac{\sigma^2}{b}, \label{eq:use-bv0}
\end{align}
where \cref{eq:use-bv0} uses $\Delta_0:=L(\bm{\theta}^0) -L^*$ and the bounded variance \cref{asp:bv} for the initial gradient estimator $g^0=\tilde{\nabla}_{b} L(\bm{\theta}^{0})$.

Then for any convergence error $\epsilon$, by choosing learning rate $\eta\leq \frac{1}{2\beta}$, minibatch sizes $b=\frac{2\sigma^2}{\epsilon^2}, b'\leq \sqrt{b}$, and probability $p=\frac{b'}{b+b'}$, we have 
\begin{align}
	\frac{1}{T}\sum_{t=0}^{T-1}\E\ns{\nabla L(\bm{\theta}^t)} 
	&\leq O\left(\frac{1}{T}\right)
	= \epsilon^2.
\end{align}
This means that after $T=O(\frac{1}{\epsilon^2})$ steps, PAGE-type algorithms can find a suitable parameter $\hat{\bm{\theta}}$ for \XVFL such that $\E\ns{\nabla L(\hat{\bm{\theta}})} \leq \epsilon^2$, where $\epsilon$ denotes the convergence error.
Compared with the SGD-type algorithms in \cref{thm:sgd}, the variace-reduced PAGE-type algorithms in \cref{thm:page} significantly decreases the total number of communication rounds, i.e., from $T=O(\frac{1}{\epsilon^4})$ to $T=O(\frac{1}{\epsilon^2})$ by a factor of $\frac{1}{\epsilon^2}$.
\qedb

\section{Experimental Details}
\label{sec:experiment details}

We now provide more details of the experimental setup in this appendix. 
We conduct the experiments on real-world datasets, including three image datasets: \dataset{CIFAR-10}~\citep{CIFAR10}, \dataset{TinyImageNet} (restricted to five classes)~\citep{ImageNet}, and \dataset{UTKFace}~\citep{utkface}, as well as three tabular datasets: \dataset{MIMIC-III}~\citep{MIMIC}, \dataset{Bank}~\citep{bank}, and \dataset{Avazu}~\citep{Fedcvt}. 
For the image datasets, features are vertically partitioned into disjoint subsets and assigned to different clients. 
For the tabular datasets, continuous features are normalized to the range $[0, 1]$, while categorical features are transformed into one-hot encodings (\dataset{Bank}) or embedding-based representations (\dataset{Avazu}). 
These preprocessed features are then partitioned into disjoint subsets and distributed to the clients. The server retains the class labels of the datasets.  

To simulate real-world scenarios where clients have partially missing features, we mask a proportion of features in the non-aligned training and test samples. 
Various feature missing rates $R_\text{miss}=\{0, 0.1, 0.3, 0.5, 0.7, 0.9, 1\}$ are applied to to thoroughly assess the impact of missing features. 

For the image datasets (\dataset{CIFAR-10}, \dataset{TinyImageNet}, and \dataset{UTKFace}), the client's local (bottom) model is implemented using a residual network with varying depths, characterized by different numbers of residual blocks. 
The corresponding server (top) model is a six-layer fully connected neural network (FCNN). 
For the tabular datasets (\dataset{MIMIC-III}, \dataset{Bank}, and \dataset{Avazu}), both the client and server models are implemented as three-layer FCNNs. 
The input dimensions of the server models were set to 2048 for image datasets and 512 for tabular datasets to accommodate differences in feature complexity and representation requirements.
The batch size is set to 50. 
The hyperparameters $\lambda_1$ and $\lambda_2$ defined in the overall loss function (see \cref{eq:overall}) are selected from the ranges listed in \cref{tab:lambda_values}.

\begin{table}[h]
	\centering
	\caption{Hyperparameter ranges for $\lambda_1$ and $\lambda_2$}
	\label{tab:lambda_values}
	\renewcommand{\arraystretch}{1.1}  
	\small
	 \resizebox{0.99\linewidth}{!}{
		\begin{tabular}{lcc}
			\toprule
			\textbf{Dataset} & \(\bm{\lambda_1}\) & \(\bm{\lambda_2}\) \\
			\midrule
			\dataset{CIFAR-10} & \{0.01, 0.02, 0.05, 0.1, 0.2, 0.5\} & \{1$\times 10^{-5}$, 2$\times 10^{-5}$, 5$\times 10^{-5}$, 1$\times 10^{-4}$, 2$\times 10^{-4}$, 5$\times 10^{-4}$\} \\
			\dataset{TinyImageNet} & \{1, 2, 5, 10, 20\} & \{5$\times 10^{-4}$, 1$\times 10^{-3}$, 2$\times 10^{-3}$, 5$\times 10^{-3}$\} \\
			\dataset{UTKFace} & \{1, 2, 5, 10, 20\} & \{5$\times 10^{-4}$, 1$\times 10^{-3}$, 2$\times 10^{-3}$, 5$\times 10^{-3}$\} \\
			\dataset{MIMIC-III} & \{5$\times 10^{-4}$, 1$\times 10^{-3}$, 2$\times 10^{-3}$, 5$\times 10^{-3}$, 1$\times 10^{-2}$\} & \{0.02, 0.05, 0.1\} \\
			\dataset{Bank} & \{1$\times 10^{-4}$, 2$\times 10^{-4}$, 5$\times 10^{-4}$, 1$\times 10^{-3}$\} & \{5$\times 10^{-5}$, 1$\times 10^{-4}$, 2$\times 10^{-4}$, 5$\times 10^{-4}$\} \\
			\dataset{Avazu} & \{1$\times 10^{-4}$, 2$\times 10^{-4}$, 5$\times 10^{-4}$, 1$\times 10^{-3}$\} & \{1$\times 10^{-5}$, 2$\times 10^{-5}$, 5$\times 10^{-5}$, 1$\times 10^{-4}$, 2$\times 10^{-4}$\} \\
			\bottomrule
		\end{tabular}
		 }
\end{table}

Vanilla Standalone refers to an independently trained model in which each client uses only its own local features for training and inference. 
It servers as a baseline for evaluating the capability of independent inference.
Vanilla VFL enables collaborative training by concatenating local feature representations from all clients at a central server. During inference, these concatenated representations is performed to generate predictions. This approach serves as a baseline for evaluating the capability of collaborative inference.
For other baselines, although previous studies typically either cannot handle missing features or do not support locally independent inference, we identify two related works, IAVFL~\citep{IAVFL} and FedCVT~\citep{Fedcvt}, that are relevant for comparison, as discussed in \cref{sec:preliminary}. 
Concretely, we compare our \XVFL with these four baselines for both independent Inference and collaborative inference modes, using test classification accuracy as the performance metric. 
All results are averaged over five random seeds to ensure robustness.

\section{Additional Experimental Results for \dataset{TinyImageNet} and \dataset{Bank}}
\label{sec:2 Clients Supplementary Details}

\begin{figure}[h]
	\centering
	\begin{subfigure}{0.48\textwidth}
		\centering
		\includegraphics[width=\linewidth]{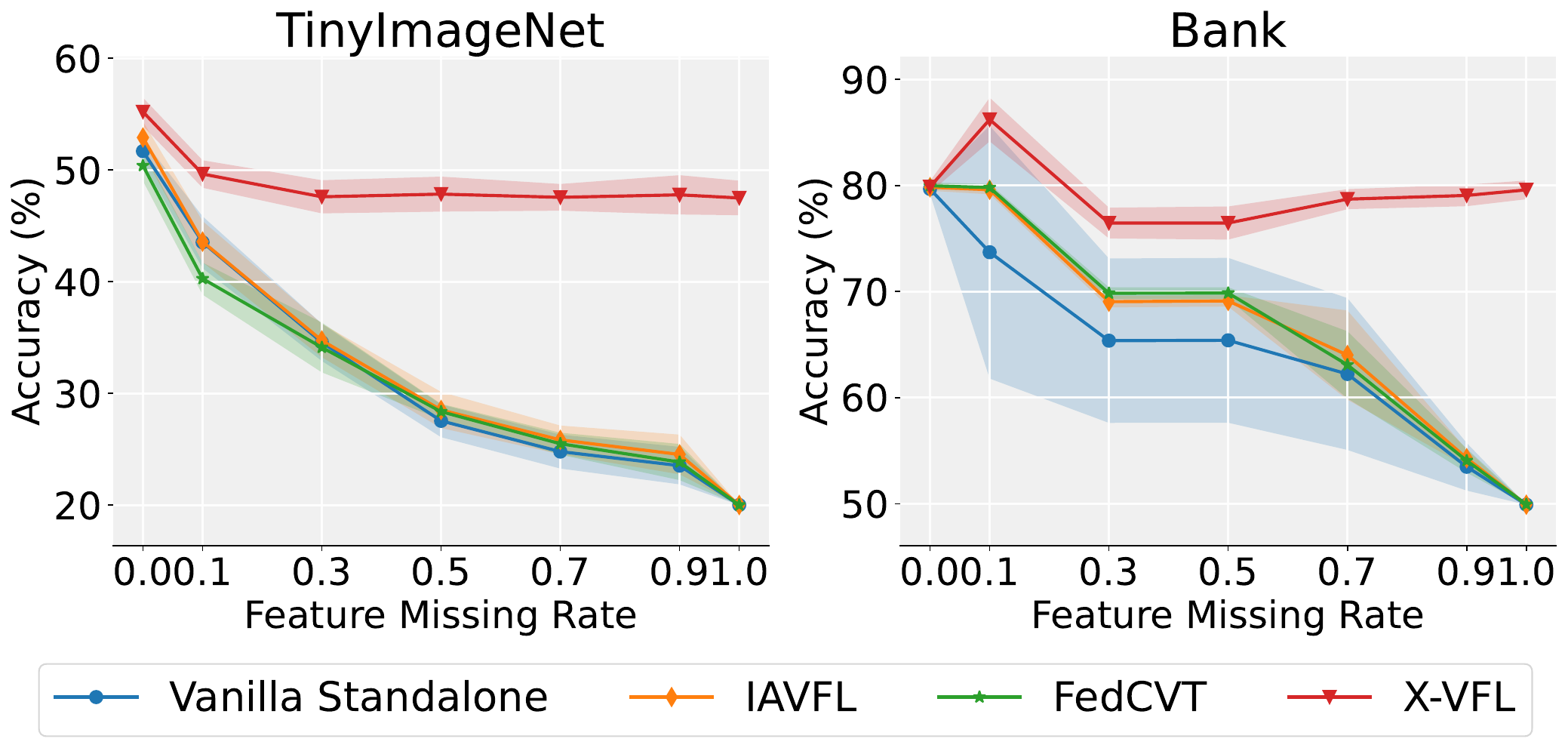}
		\caption{Independent inference mode}
		\label{fig:more_performance_ind_missing_features}
	\end{subfigure}\hspace{4mm}
	\begin{subfigure}{0.48\textwidth}
		\centering
		\includegraphics[width=\linewidth]{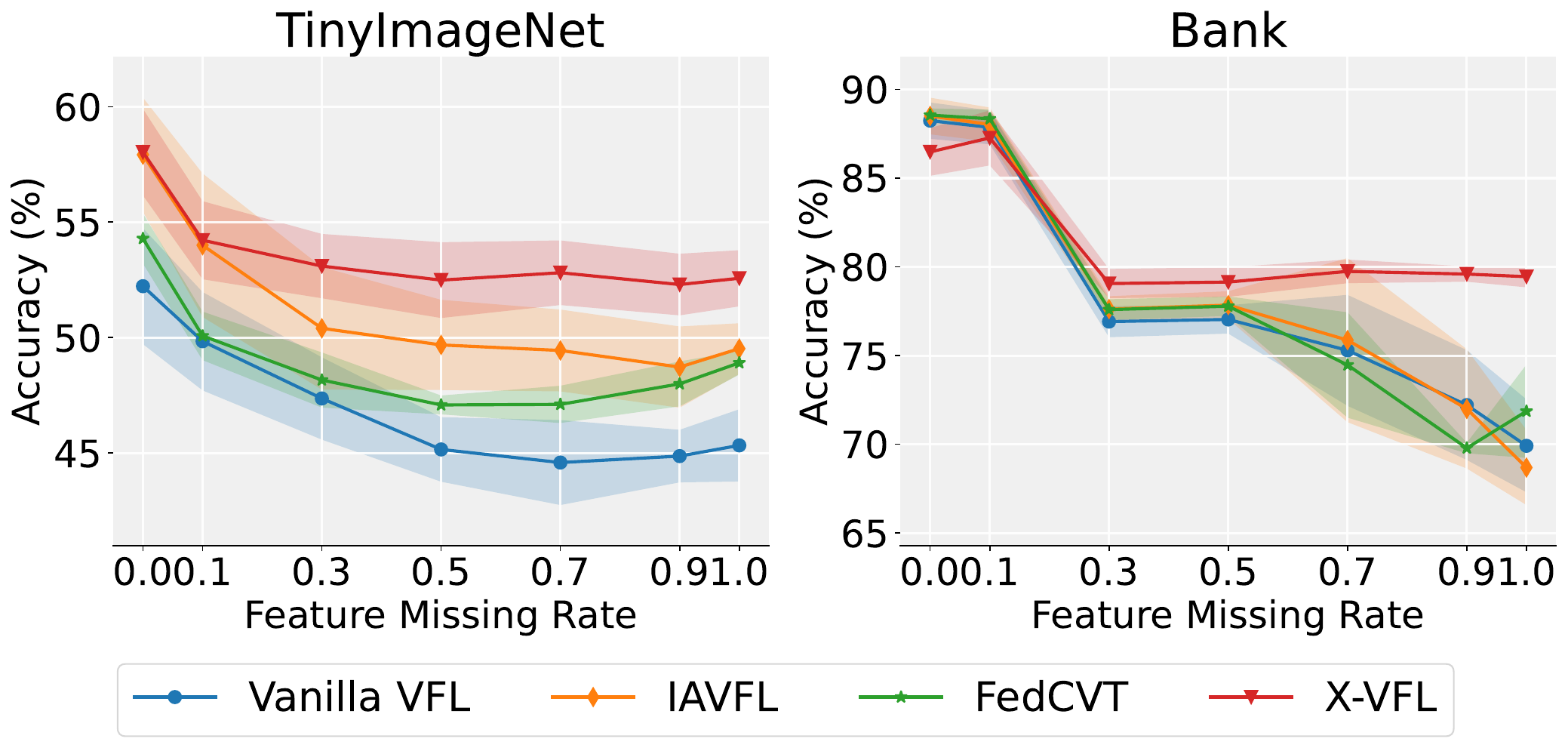}
		\caption{Collaborative inference mode}
		\label{fig:more_performance_col_missing_features}
	\end{subfigure}
	\caption{Performance comparison under varying feature missing rates on the \dataset{TinyImagenet} and \dataset{Bank} datasets.}
	\label{fig:more_performance_missing_features_combined}
\end{figure}

\begin{figure}[h]
	\centering
	\includegraphics[width =0.5\linewidth]{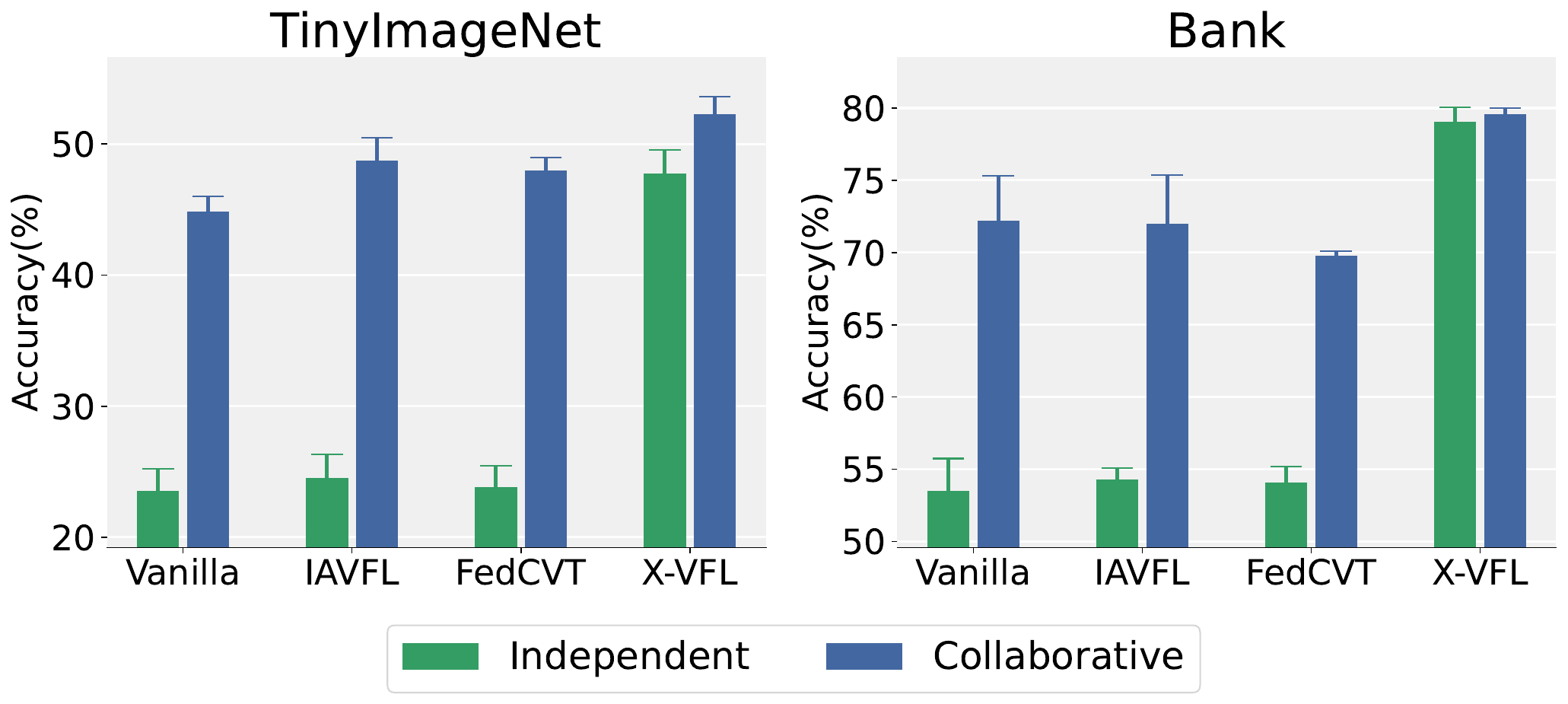}
	\caption{Performance difference between independent and collaborative inference modes under feature missing rate $R_\text{miss}=0.9$ on the \dataset{TinyImagenet} and \dataset{Bank} datasets.}
	\label{fig:more_performance_diff}
\end{figure}

Similar to \cref{subsec:Results_missing features}, this appendix provides additional experimental results on the \texttt{TinyImageNet} and \dataset{Bank} datasets under varing feature missing rates.
The results are presented in \cref{fig:more_performance_missing_features_combined} and \cref{fig:more_performance_diff}. 
Moreover, the results demonstrate the same trend: our \XVFL significantly outperforms the baselines, particularly in the independent inference mode---one of the key challenges that \XVFL is designed to address.

\section{General Multiple Clients Setting}\label{sec:general clients}

In this appendix, we present the extension of our \XVFL framework to the general case with multiple $k$ clients. 
This extension enhances the practicality of \XVFL and offers a more scalable and effective approach for broader applications.
The framework formulation and corresponding experimental results are provided in \cref{subsec:kclients} and \cref{subsec:kclients experiments}, respectively.

\subsection{Formulation for multiple clients}\label{subsec:kclients}
Similar to \cref{subsec:Framework Description}, let $f_{1}, \dots, f_{k}$ denote the local models for the $k$ clients. Then $\bm{E}_{i} = f_i(\bm{x_i})$ denotes the feature embeddings/representations for client $i$, $i \in [k]$.
For the general case of $k$ clients, the overall loss function can be formulated as the same one as in \cref{eq:overall}:
\begin{equation}\label{eq:overall_k}
	L = L_{\text{decision}} + \lambda_1 L_{\text{DSAlign}_{1}} + \lambda_2 L_{\text{DSAlign}_{2}}.
\end{equation}

Compared with two-client setting, now the classification decision loss for $k$ clients is formulated as follows:
\begin{equation}\label{eq:l_ce_aligned_non_aligned_k} 
	L_{\text{decision}} = \sum_{i \in \mathcal{M}}\ell(h(\bm{E}_{i}), y) + \ell(h(\frac{\bm{E}^{\prime}}{m}), y)  + \ell(h(\frac{\tilde{\bm{E}^{\prime}}+\bm{E}^{\prime}}{k}), y) + \sum_{\substack{i=1}}^k\ell(h(\frac{\tilde{\bm{E}_{i}}+\bm{E}_{-i}}{k}), y),
\end{equation}
where 
$\bm{E}^{\prime} = \sum_{j \in \mathcal{M}} \bm{E}_j$, 
$\tilde{\bm{E}^{\prime}} = \sum_{i \in \{1, \dots, k\} \setminus \mathcal{M}} \tilde{\bm{E}}^{\prime}_{i}$,
$\mathcal{M} \subseteq \{1, \dots, k\}$ denotes the subset of clients that hold full local features (i.e., clients without missing features), where $|\mathcal{M}|=m$ for some $m \leq k$, 
$\bm{E}_{-i} = \sum_{j\in\{1, \ldots, k\} \setminus \{i\}}  \bm{E}_j$, and 
\begin{equation}\label{eq:vfl_xcom_k_1} 
	\tilde{\bm{E}}_{i} = f_{i}(\tilde{\bm{X}}_{i}), 
	\quad \tilde{\bm{X}}_{i} = \XCom_{i}(\frac{\bm{E}_{-i}}{k-1}),  \quad 1 \leq i \leq k,
\end{equation}
\begin{equation}\label{eq:vfl_xcom_k_2} 
	\tilde{\bm{E}}^{\prime}_{i} = f_{i}(\tilde{\bm{X}}^{\prime}_{i}),
	\quad \tilde{\bm{X}}^{\prime}_{i} = \XCom_{i}(\frac{\bm{E}^{\prime}}{m}), \quad 1 \leq i \leq k.
\end{equation}

For aligned data, the following terms in \cref{eq:l_ce_aligned_non_aligned_k} are activated: $\sum_{i \in \mathcal{M}}\ell(h(\bm{E}_{i}), y)$, $\ell(h(\frac{\bm{E}^{\prime}}{m}), y)$, and $\sum_{\substack{i=1}}^k\ell(h(\frac{\tilde{\bm{E}_{i}}+\bm{E}_{-i}}{k}), y)$.
The term $\ell(h(\frac{\tilde{\bm{E}^{\prime}}+\bm{E}^{\prime}}{k}), y)$ is not activated, as $\tilde{\bm{E}^{\prime}}$ is empty when $\mathcal{M}$ includes all clients (i.e., $k=m$).
For non-aligned data, the actived terms are $\sum_{i \in \mathcal{M}}\ell(h(\bm{E}_{i}), y)$, $\ell(h(\frac{\bm{E}^{\prime}}{m}), y)$, 
and $\ell(h(\frac{\tilde{\bm{E}^{\prime}}+\bm{E}^{\prime}}{k}), y)$.

The last two loss components introduced by our \DSAlign module in the general case of $k$ clients are formulated as follows:
\begin{equation}\label{eq:DSAlign_1_k}
	L_{\text{DSAlign}_{1}} = \sum_{\substack{i=1}}^k\ell(h(\tilde{\bm{E}}_{i}), h(\bm{E}_{i}) ),
\end{equation}
and
\begin{equation}\label{eq:DSAlign_2_k}
	L_{\text{DSAlign}_{2}} = \sum_{\substack{i=1}}^k\ell( h(\bm{E}_{i}), h(\frac{\sum_{\substack{i=1}}^k\bm{E}_{i}}{k})).
\end{equation}

\subsection{Experimental results for multiple clients}
\label{subsec:kclients experiments}

In this appendix, we provide the experimental results for the general multiple clients setting.
We use \dataset{CIFAR-10}~\citep{CIFAR10} as the image dataset and \dataset{MIMIC-III}~\citep{MIMIC} as the tabular dataset. 
In particular, we conduct  experiments with $k=4$ clients.
Regarding the baselines, since FedCVT~\citep{Fedcvt} is specifically designed for the two-client scenario, it cannot be extended to the multi-client setting.
Therefore, we include three baseline methods (Vanilla Standalone, Vanilla VFL, IAVFL) in the multi-client experiments, consistent with those used in the two-client setting, excluding only FedCVT.
The implementation details remain the same as in the two-client case (see \cref{sec:experiment details}), except that the batch size is set to 100. 
In the multi-client setting, we simply set the hyperparameters $\lambda_1=\lambda_2= 0.01$. 

\begin{figure}[h]
	\centering
	\begin{subfigure}{0.48\textwidth}
		\centering
		\includegraphics[width=\linewidth]{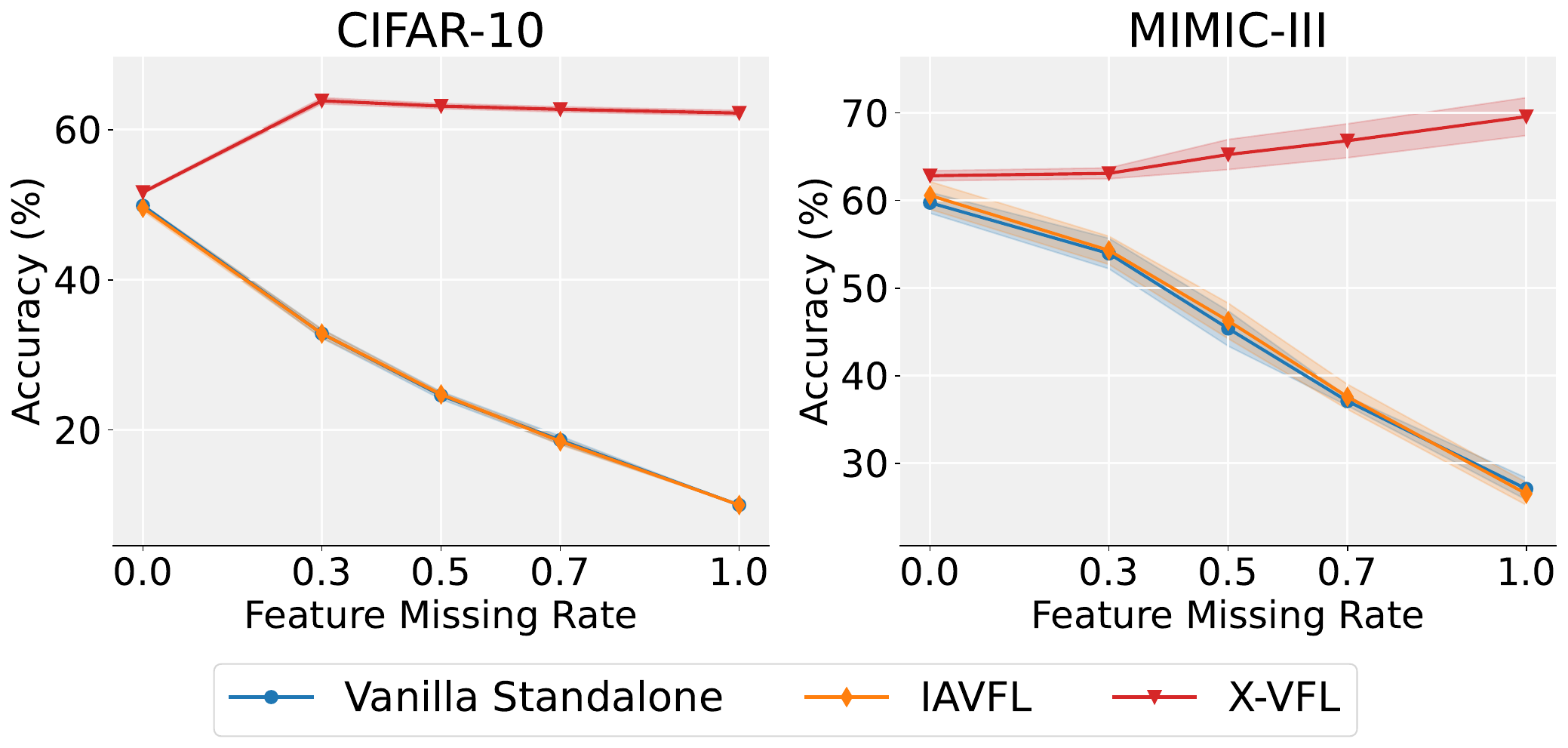}
		\caption{Independent inference mode}
		\label{fig:line_chart_row1_4clients}
	\end{subfigure}\hspace{4mm}
	\begin{subfigure}{0.48\textwidth}
		\centering
		\includegraphics[width=\linewidth]{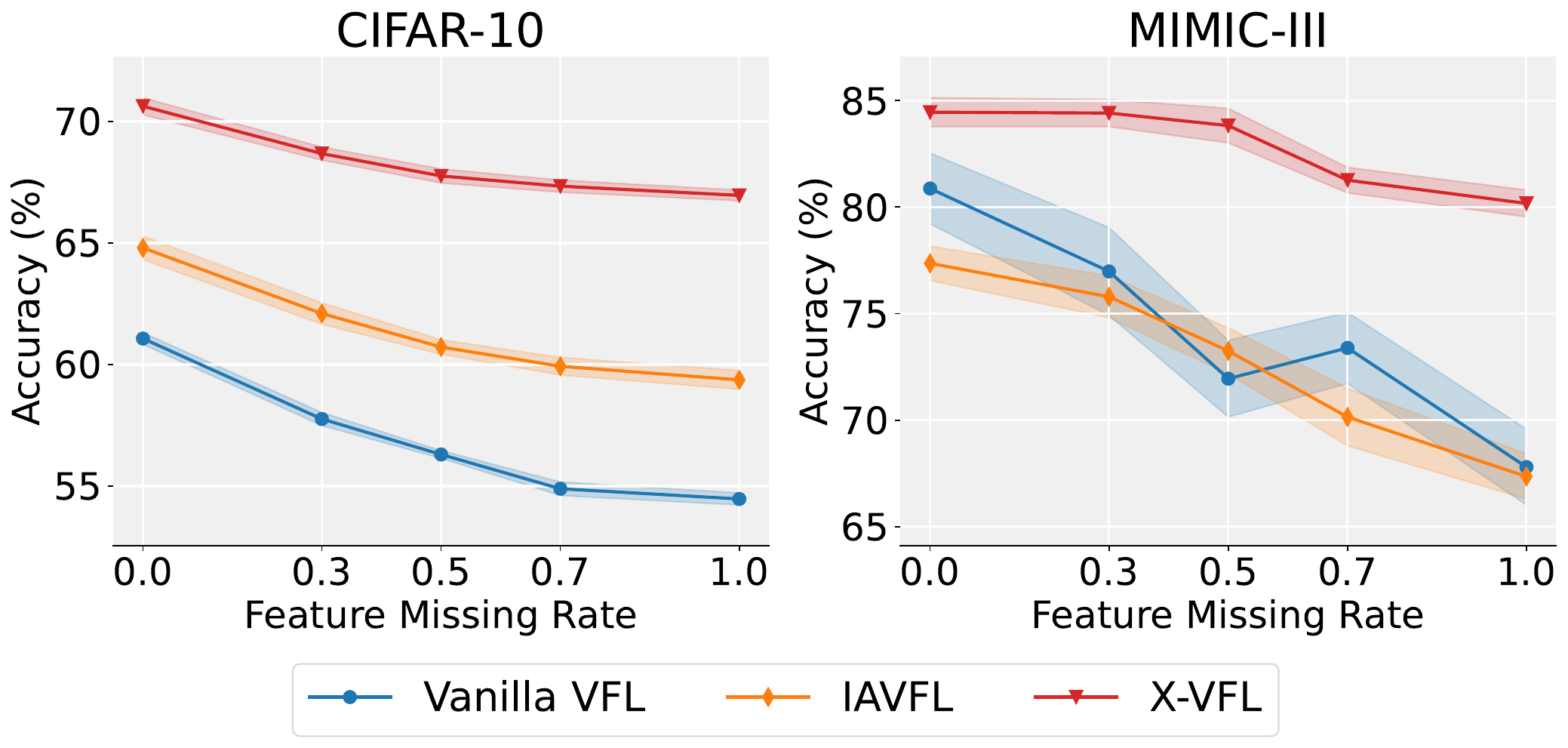}
		\caption{Collaborative inference mode}
		\label{fig:line_chart_row2_4clients}
	\end{subfigure}
	\caption{Performance comparison under varying feature missing rates on the \dataset{CIFAR-10} and \dataset{MIMIC-III} datasets in the multiple clients setting.}
	\label{fig:line_chart_4clients}
%
%
	\vspace{5mm}
	\centering
	\includegraphics[width = 0.5\linewidth]{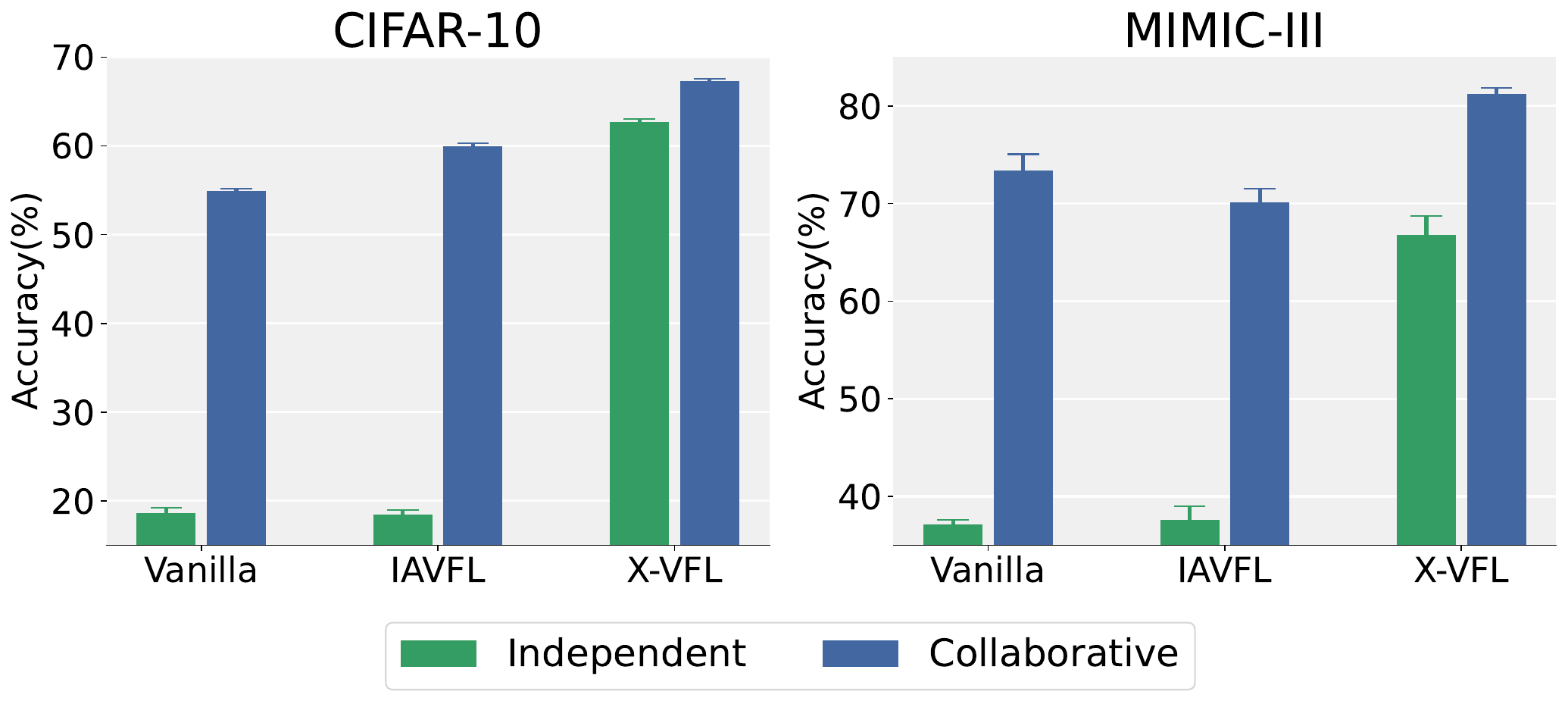}
	\caption{Performance comparison between independent and collaborative inference modes under feature missing rate of $0.7$ on the \dataset{CIFAR-10} and \dataset{MIMIC-III} datasets in the multiple clients setting.}
	\label{fig:bar_1_4_clients}
%
%
	\vspace{5mm}
	\centering
	\begin{subfigure}{0.48\textwidth}
		\centering
		\includegraphics[width=\linewidth]{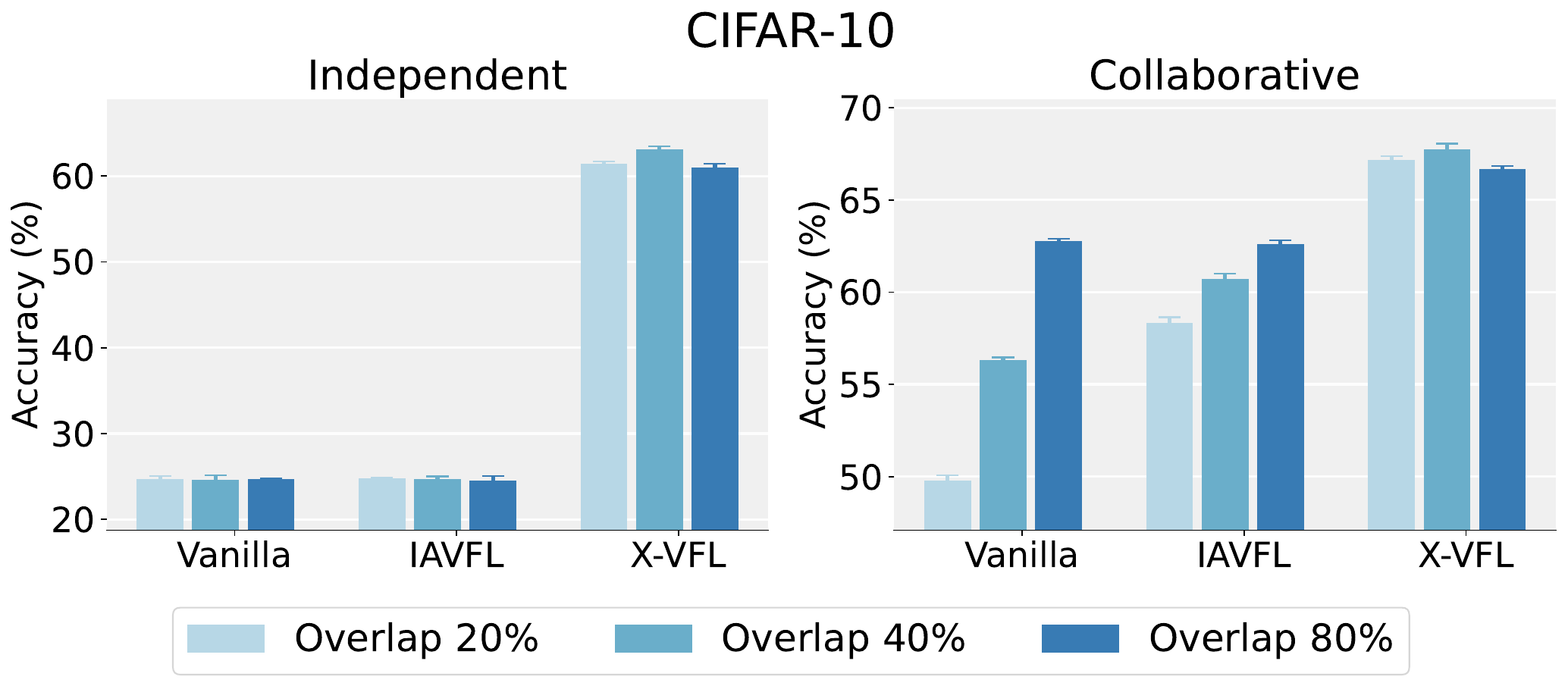}
		\caption{\dataset{CIFAR-10} dataset}
		\label{fig:difOverlap_cifar10_4clients}
	\end{subfigure}\hspace{4mm}
	\begin{subfigure}{0.48\textwidth}
		\centering
		\includegraphics[width=\linewidth]{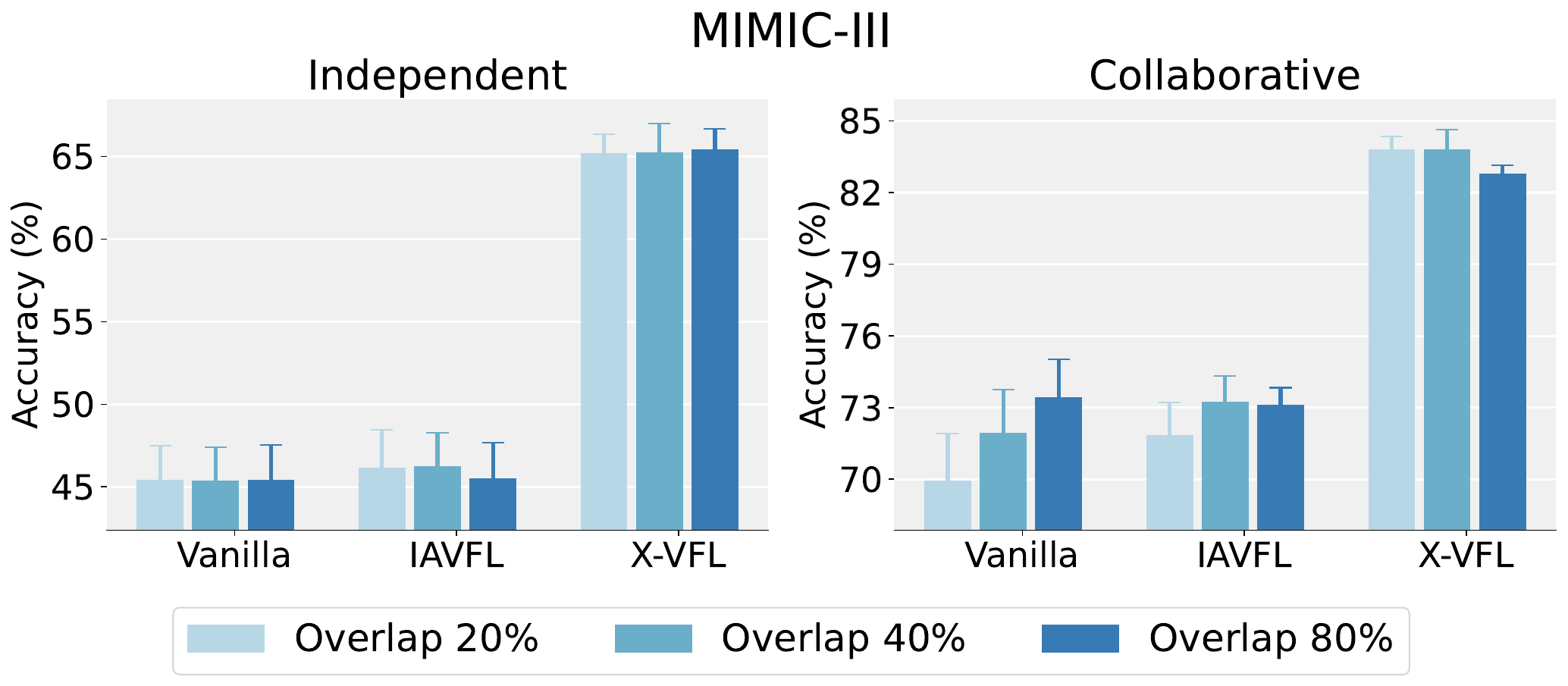}
		\caption{\dataset{MIMIC-III} dataset}
		\label{fig:difOverlap_mimic_4clients}
	\end{subfigure}
	\caption{Performance comparison under varying overlap ratios on the \dataset{CIFAR-10} and \dataset{MIMIC-III} datasets in the multiple clients setting.}
	\label{fig:bar_chart_2_cifar10_mimic-4}
	\vspace{-5mm}
\end{figure}

\cref{fig:line_chart_4clients} presents the results under varying feature missing rates, $R_\text{miss}=\{0, 0.3, 0.5, 0.7, 1.0\}$, for both independent and collaborative inference modes. A comparison of performance between these two modes at a missing rate $R_\text{miss}=0.7$ is shown in \cref{fig:bar_1_4_clients}. 
The results follow the same trend observed in the two-client case: our \XVFL significantly outperforms the baselines, particularly in the independent inference mode---one of the key challenges that \XVFL is designed to address.
Moreover, \cref{fig:bar_1_4_clients} highlights the advantage of \XVFL in effectively narrowing the accuracy gap between independent and collaborative inference modes.

We also evaluate the effect of varying overlap ratios of aligned data samples, using 20\%, 40\%, and 80\% aligned samples, as shown in \cref{fig:bar_chart_2_cifar10_mimic-4}.
Similar to the two-client case, \XVFL consistently outperforms the baseline methods across all overlap ratios in both independent and collaborative inference modes.

\end{document}